%% file: main.tex
\documentclass[10pt,twocolumn,letterpaper]{article}

\usepackage{cvpr}              

\usepackage[acronym,nogroupskip,nonumberlist,nopostdot,nohypertypes={acronym,notation}]{glossaries}
\makeglossaries
\input{preamble}
\usepackage[accsupp]{axessibility} 

\definecolor{cvprblue}{rgb}{0.21,0.49,0.74}
\usepackage[pagebackref,breaklinks,colorlinks,citecolor=cvprblue]{hyperref}
\usepackage{tikz}
\usepackage{afterpage}
\usepackage{comment}

\def\paperID{4665} 
\def\confName{CVPR}
\def\confYear{2024}

\definecolor{lionBlue}{HTML}{2274A5}
\definecolor{lionY}{HTML}{D3A40A}
\definecolor{lionP}{HTML}{544D71}

\title{NeRF Director: Revisiting View Selection in Neural Volume Rendering}

\author{Wenhui Xiao\textsuperscript{1,2}, 
Rodrigo Santa Cruz\textsuperscript{1,2}, 
David Ahmedt-Aristizabal\textsuperscript{1,2}, \\
Olivier Salvado\textsuperscript{1,2}, 
Clinton Fookes\textsuperscript{1}, 
Leo Lebrat\textsuperscript{1,2} \\
Queensland University of Technology\textsuperscript{1}, 
CSIRO Data61\textsuperscript{2} \\
 {\tt\small wenhui.xiao@hdr.qut.edu.au, \{rodrigo.santacruz, leo.lebrat\}@csiro.au}\\
 \small \url{https://wenwhx.github.io/nerfdirector}
}

\begin{document}


\twocolumn[{%
\renewcommand\twocolumn[1][]{#1}%
\maketitle
\begin{center}
    \centering
    \captionsetup{type=figure}
    \input{fig/fig1}
\end{center}
}]

\input{sec/0_abstract} 
\vspace{-8pt}
\input{sec/1_intro}

\input{sec/2_related_works}

\input{sec/3_motivation}
\input{sec/4_method}

\input{sec/5_exp}
\input{sec/6_summary}

\input{sec/X_suppl}

\end{document}

%% file: preamble.tex
\usepackage{stmaryrd}
\usepackage{algpseudocode}
\usepackage[ruled,linesnumbered,vlined,noend]{algorithm2e}
\usepackage{multirow}
\usepackage[table]{xcolor} 
\usepackage[sectionbib]{chapterbib}

\DeclareMathOperator*{\argmax}{arg\,max}

\newacronym{snr}{SNR}{signal-to-noise ratio}
\newacronym{sinr}{SINR}{signal-to-interference-plus-noise ratio}
\newacronym{sir}{SIR}{signal-to-interference ratio}
\newacronym{sqr}{SQR}{signal-to-quantization-noise ratio}
\newacronym{sqnr}{SQNR}{signal-to-quantization-plus-noise ratio}

\newacronym{ber}{BER}{bit error rate}
\newacronym{evm}{EVM}{error vector magnitude}
\newacronym{isi}{ISI}{intersymbol interference}

\newacronym{bfsk}{BFSK}{binary frequency shift keying}
\newacronym{qam}{QAM}{quadrature amplitude modulation}
\newacronym{mqam}{MQAM}{M-ary quadrature amplitude modulation}
\newacronym{dsss}{DSSS}{direct-sequence spread spectrum}
\newacronym{ofdm}{OFDM}{orthogonal frequency-division multiplexing}
\newacronym{ofdma}{OFDMA}{orthogonal frequency-division multiple access}
\newacronym{fdd}{FDD}{frequency-division duplexing}
\newacronym{tdd}{TDD}{time-division duplexing}
\newacronym{fdma}{FDMA}{frequency-division multiple access}
\newacronym{tdma}{TDMA}{time-division multiple access}
\newacronym{sdma}{SDMA}{space-division multiple access}

\newacronym{ls}{LS}{least-squares}
\newacronym{lms}{LMS}{least mean squares}
\newacronym{omp}{OMP}{orthogonal matching pursuit}

\newacronym{zf}{ZF}{zero-forcing}
\newacronym{mmse}{MMSE}{minimum mean square error}
\newacronym{mse}{MSE}{mean square error}

\newacronym{fft}{FFT}{fast Fourier transform}
\newacronym{dft}{DFT}{discrete Fourier transform}
\newacronym{dtft}{DTFT}{discrete-time Fourier transform}
\newacronym{ctft}{CTFT}{continuous-time Fourier transform}

\newacronym{adc}{ADC}{analog-to-digital converter}
\newacronym{dac}{DAC}{digital-to-analog converter}
\newacronym{fpga}{FPGA}{field-programmable gate array}
\newacronym{enob}{ENOB}{effective number of bits}

\newacronym{rv}{r.v.}{random variable}
\newacronym{svd}{SVD}{singular value decomposition}
\newacronym{sdp}{SDP}{semidefinite programming}
\newacronym{psd}{PSD}{positive semidefinite}
\newacronym{nsd}{NSD}{negative semidefinite}

\newacronym{agc}{AGC}{automatic gain control}
\newacronym{rf}{RF}{radio frequency}
\newacronym{los}{LOS}{line-of-sight}
\newacronym{nlos}{NLOS}{non-line-of-sight}
\newacronym{ple}{PLE}{path loss exponent}
\newacronym[plural=dB,firstplural=decibels (dB)]{db}{dB}{decibel}
\newacronym[plural=dBm,firstplural=decibel milliwatts (dBm)]{dbm}{dBm}{decibel milliwatts}
\newacronym{pa}{PA}{power amplifier}
\newacronym{lna}{LNA}{low noise amplifier}
\newacronym{cw}{CW}{continuous wave}
\newacronym{papr}{PAPR}{peak-to-average power ratio}
\newacronym{tx}{TX}{transmitter}
\newacronym{rx}{RX}{receiver}
\newacronym{sdr}{SDR}{software-defined radio}
\newacronym{usrp}{USRP}{Universal Software Radio Peripheral}
\newacronym{lo}{LO}{local oscillator}
\newacronym{mmwave}{mmWave}{millimeter-wave}
\newacronym{eirp}{EIRP}{effective isotropic radiated power}

\newacronym{csma}{CSMA}{carrier-sense multiple access}
\newacronym{csmaca}{CSMA/CA}{carrier-sense multiple access with collision avoidance}
\newacronym{csmacd}{CSMA/CD}{carrier-sense multiple access with collision detection}
\newacronym{mac}{MAC}{medium access control}
\newacronym{phy}{PHY}{physical layer}

\newacronym{4g}{4G}{fourth generation}
\newacronym{lte}{LTE}{Long-Term Evolution}
\newacronym{5g}{5G}{fifth generation}
\newacronym{nr}{NR}{New Radio}
\newacronym{5gnr}{5G NR}{5G New Radio}
\newacronym{ieee}{IEEE}{Institute of Electrical and Electronics Engineers}
\newacronym{lan}{LAN}{local area network}
\newacronym{wlan}{WLAN}{wireless local area network}
\newacronym{bs}{BS}{base station}
\newacronym{ue}{UE}{user equipment}
\newacronym{ul}{UL}{uplink}
\newacronym{dl}{DL}{downlink}
\newacronym{qos}{QoS}{Quality of Service}
\newacronym{fcc}{FCC}{Federal Communications Commission}
\newacronym{iab}{IAB}{integrated access and backhaul}
\newacronym{hetnet}{HetNet}{heterogeneous network}

\newacronym{siso}{SISO}{single-input single-output}
\newacronym{mimo}{MIMO}{multiple-input multiple-output}
\newacronym{sumimo}{SU-MIMO}{single-user \gls{mimo}}
\newacronym{mumimo}{MU-MIMO}{multi-user \gls{mimo}}
\newacronym{ula}{ULA}{uniform linear array}
\newacronym[\glslongpluralkey={angles of arrival}]{aoa}{AoA}{angle of arrival}
\newacronym[\glslongpluralkey={angles of departure}]{aod}{AoD}{angle of departure}
\newacronym{dof}{DoF}{degrees of freedom}
\newacronym{csi}{CSI}{channel state information}
\newacronym{csit}{CSIT}{\gls{csi} at the transmitter}
\newacronym{csir}{CSIR}{\gls{csi} at the receiver}
\newacronym{cs}{CS}{compressed sensing}

\newacronym{elf}{ELF}{extremely low frequency}
\newacronym{slf}{SLF}{super low frequency}
\newacronym{ulf}{ULF}{ultra low frequency}
\newacronym{vlf}{VLF}{very low frequency}
\newacronym{lf}{LF}{low frequency}
\newacronym{mf}{MF}{medium frequency}
\newacronym{hf}{HF}{high frequency}
\newacronym{vhf}{VHF}{very high frequency}
\newacronym{uhf}{UHF}{ultra high frequency}
\newacronym{shf}{SHF}{super high frequency}
\newacronym{ehf}{EHF}{extremely high frequency}
\newacronym{thf}{THF}{tremendously high frequency}

\newacronym{2d}{2D}{two-dimensional}
\newacronym{3d}{3D}{three-dimensional}

\newacronym{nerf}{NeRF}{Neural Radiance Field}
\newacronym{inr}{INR}{implicit neural representation}
\newacronym{sdf}{SDF}{Signed Distance Function}
\newacronym{deepsdf}{DeepSDF}{Deep Signed Distance Function}

\newacronym{fps}{FPS}{farthest point sampling}
\newacronym{fvs}{FVS}{farthest view sampling}
\newacronym{rs}{RS}{random sampling}
\newacronym{hs}{IGS}{information gain-based sampling}
\newacronym{vmf}{vMF}{von Mises-Fisher}

\newacronym{mlp}{MLP}{Multilayer Perceptron}
\newacronym{pe}{PE}{Positional Encoding}
\newacronym{sfm}{SfM}{structure-from-motion}
\newacronym{mvs}{MVS}{multi-view stereo}
\newacronym{vit}{ViT}{Vision Transformer}
\newacronym{slam}{SLAM}{Simultaneous Localization and Mapping}
\newacronym{gan}{GAN}{Generative Adversarial Network}
\newacronym{brdf}{BRDF}{Bidirectional Reflectance Distribution function}

\newacronym{sota}{SOTA}{state-of-the-art}

\newacronym{cad}{CAD}{Computer-aided design}

\newacronym{psnr}{PSNR}{peak signal-to-noise ratio}
\newacronym{ssim}{SSIM}{structural similarity index measure}

%% file: fig/fig1.tex
\vspace*{-1.8em}\begin{subfigure}[]{0.36\textwidth}
        \begin{tikzpicture}
        \node[inner sep=0pt] (tikzmagical) at (0,0) 
        {\includegraphics[width=0.9\textwidth]{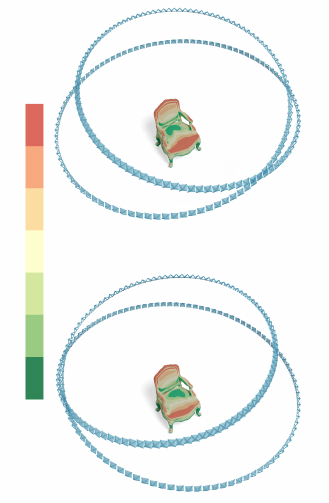}};
        \node[rotate=90, anchor=center,scale=1.16] at (-3,-0.0) {View coverage};
        \node[anchor=center,scale=1.1] at (-2.2,3.2) {high};
        \node[anchor=center,scale=1.1] at (-2.2,-3.20) {low};
        \node[anchor=west, text=lionBlue,scale=1.1] at (1.76,-4.36) {\textbf{Rot} $90^\circ$};
        \node[anchor=west, text=lionBlue,scale=1] at (1.76,4.36) {\textbf{Original}};
        \end{tikzpicture}
        \caption{}
        \label{fig:motivation-a}
  \end{subfigure}
  \hfill
  \begin{minipage}{0.58\textwidth}
      \begin{subfigure}[]{\textwidth}
        \centering
        \includegraphics[width=0.936\textwidth]{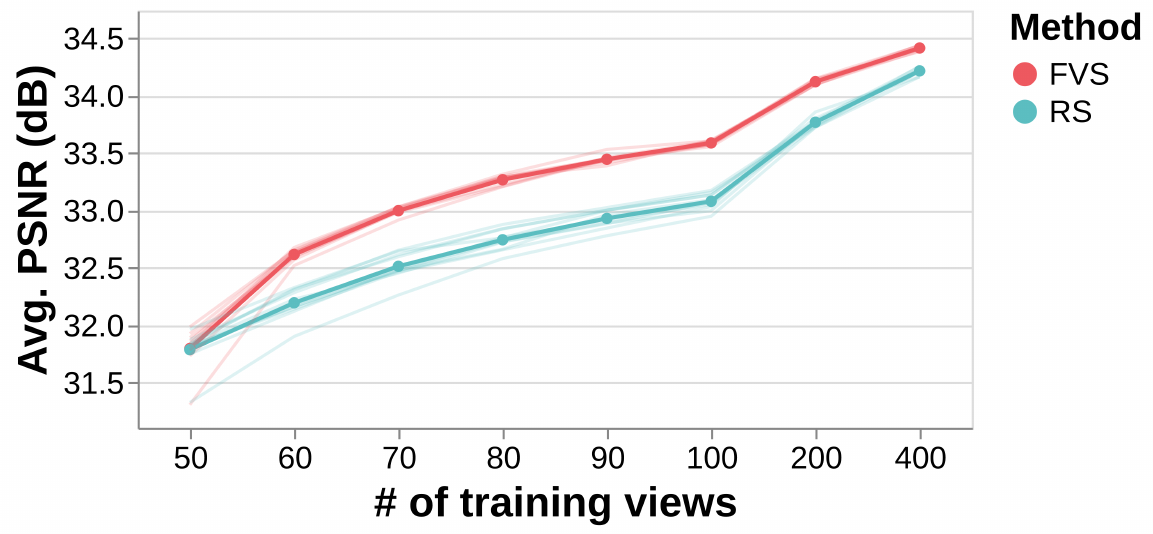}
        \caption{}
        \label{fig:motivation-b}
      \end{subfigure}
      \hfill
      \begin{subfigure}[]{\textwidth}
        \begin{tikzpicture}
          \node[inner sep=0pt] (tikzmagical) at (0,0) 
          {\includegraphics[width=\textwidth]{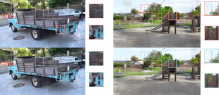}};
          \node[rotate=90, anchor=center,scale=1.10] at (5.28,1.13) {RS {\scriptsize (110 v)}};
          \node[rotate=90, anchor=center,scale=1.10] at (5.28,-1.13) {FVS {\scriptsize (80 v)}};
        \end{tikzpicture}
        \caption{}
        \label{fig:motivation-c}
      \end{subfigure}      
  \end{minipage}
  \caption{Overview of NeRF Director. \textbf{(a)} Different test camera selections result in different error measurements on the target object, which ultimately can result in SOTA ranking inversion. \textbf{(b)} In a synthetic setting and for InstantNGP~\cite{Muller2022InstantNGP}, a better view selection algorithm such as \acrfull{fvs} can reach 33 dB of PSNR using only 70 training views, whereas the traditional \acrfull{rs} would require 150 views to achieve similar performances. \textbf{(c)} With fewer views, our view selection method outperforms traditional random view selection.
  }
  \label{fig:motivation}




%% file: sec/0_abstract.tex
\begin{abstract}
\vspace{-12pt}

Neural Rendering representations have significantly contributed to the field of 3D computer vision. Given their potential, considerable efforts have been invested to improve their performance. Nonetheless, the essential question of selecting training views is yet to be thoroughly investigated. This key aspect plays a vital role in achieving high-quality results and aligns with the well-known tenet of deep learning: ``garbage in, garbage out''. 
In this paper, we first illustrate the importance of view selection by demonstrating how a simple rotation of the test views within the most pervasive NeRF dataset can lead to consequential shifts in the performance rankings of state-of-the-art techniques. 
To address this challenge, we introduce a unified framework for view selection methods and devise a thorough benchmark to assess its impact.
Significant improvements can be achieved without leveraging error or uncertainty estimation but focusing on uniform view coverage of the reconstructed object, resulting in a training-free approach.
Using this technique, we show that high-quality renderings can be achieved faster by using fewer views. We conduct extensive experiments on both synthetic datasets and realistic data to demonstrate the effectiveness of our proposed method compared with random, conventional error-based, and uncertainty-guided view selection.

\end{abstract}

%% file: sec/1_intro.tex
\section{Introduction}
\label{sec:intro}

\glspl{nerf}~\cite{Mildenhall2020NeRF} can effectively learn the geometry and appearance of a 3D scene and have succeeded in novel view rendering.
Ongoing research is devoted to enhancing \gls{nerf}'s quality and efficiency by improving network architectures~\cite{Barron2021MipNeRF, Chen2022TensoRF, Verbin2022RefNeRF, Fridovich2022Plenoxels, Muller2022InstantNGP}, and tackling the challenges from insufficient training views~\cite{Jain2021DietNeRF, Yu2020PixelNeRF, Zhang2022RapNeRF} or inaccurate camera poses~\cite{Lin2021BARF, Wang2021NeRFMinus, Truong2022SPARF, Chen2023L2GBarf}.

Another focus of recent efforts study how to sample 3D training primitives (e.g., views, rays, and points) to efficiently and effectively train a NeRF model.
For instance, EfficientNeRF~\cite{Hu2022efficientnerf} manages to sample valid and pivotal points based on density and accumulated transmittance.
~\citet{Zhang2023FewerRays} propose to improve the learning of the radiance field without sacrificing quality significantly by selectively shooting rays in important regions.
However, we argue that view selection should be the essential problem for data sampling. 
Views serve as \textit{the root source of points and rays}, yet their selection remains an uncharted topic needing more comprehensive exploration.

View selection is crucial for both the evaluation and training of \gls{nerf}.
On the one hand, a seamless rotation on the testing camera poses, leading to a different error measurement distribution on the reconstructed object as shown in \Cref{fig:motivation-a}, can yield an inversion in the rankings of the \gls{sota} \gls{nerf} methods.
We underscore the importance of fair error assessment, which requires giving equal importance to every part of the scene.
On the other hand, view selection also plays a pivotal role in training a \gls{nerf} model.
It can be noticed from \Cref{fig:motivation-b} that in a noise-free synthetic setting, the performances of \gls{nerf} at convergence are influenced by the sampling method used to select its training views. 
Furthermore, \Cref{fig:motivation-c} indicates that fewer, well-chosen views can yield better novel-view rendering performance.
With the ubiquity of smartphones with continually improving camera specifications, sourcing high-resolution data is no longer an issue. 
The prospect of a lightweight view selection algorithm is high as it contributes to improving the utility of deployable \gls{nerf}-based approaches.
This motivates us to study and answer the question -- \textit{what is an effective way to select a given number of views from a large amount of training data to achieve better rendering performance?}

This paper introduces a comprehensive view selection assessment framework, \gls{nerf} Director, and explores the impact of different view selection schemes. 
First, we propose a robust method for generating a test split from a set of posed images without geometrical priors.
It targets minimizing the sampling variance among testing views and empirically yields a more consistent evaluation.
Then, we investigate typical sampling methods including \gls{rs}, and two types of heuristic sampling: \gls{fvs} and \gls{hs}.
The \gls{fvs} is derived from \acrlong{fps}~\cite{Eldar1997FPS}, and selects informative and spatially distributed training views.
\Gls{hs} methods allocate candidates with information gain in terms of error or uncertainty~\cite{Pan2022ActiveNeRF, Sunderhauf2022DensityAwareEnsembles}, and each time picks the most profitable candidate. 
We also devise a variant of \gls{hs} methods leveraging Lloyd's algorithm~\cite{lloyd1982lloyd} to mitigate the over-sampling effect of greedy view selection strategies. 
Finally, we conduct comprehensive experiments on both synthetic and real-world datasets.

The experimental results show that varying the choice of view selection schemes can result in a PSNR difference of up to $1.9$ dB.
For a fixed view budget, \gls{fvs} reaches the converged quality of traditional \gls{rs} significantly faster with up to 4$\times$ speedup.
This indicates the critical impact of view selection on the performance of \gls{nerf} --- a factor that should not be overlooked when the community continually improves the \gls{sota} performance of \gls{nerf} models.
Our detailed analysis yields interesting and important findings, listed below.

\begin{itemize}
    \item The diversity within selected training views significantly contributes to the final reconstruction and should be given the highest initial consideration. 
    \item \Gls{hs}, being time-consuming, heavily depends on the accuracy of the adopted information. Placement based on noisy information may result in inferior results when compared to \gls{rs}.
    \item \Gls{hs} exhibits sensitivity to error and tend to cluster on complex regions of the scene, which may pose challenges for effective learning by \gls{nerf}. To enhance their performance, a relaxation step becomes necessary.
\end{itemize}

%% file: sec/2_related_works.tex
\section{Related works}
\label{sec:related_works}

The importance of view sampling for traditional 3D reconstruction has been a thoroughly studied topic, and it is demonstrated that the reconstruction's quality heavily depends on viewpoint selection~\cite{agarwal2009BuildingRomeInADay, jancosek2009scalablemvs, Mendez2017ScenicPathplanner}. Nonetheless, this topic has not been extensively explored for NeRFs.

This paper focuses on exploring novel approaches to select training views from an extensive training pool so as to achieve optimal rendering quality.
This section discusses existing work on training data sampling, which can be categorized into three types: uncertainty-guided, error-guided, and scene coverage-based methods.

\noindent\textbf{Uncertainty-Guided Methods:}
Uncertainty estimation in neural networks plays a crucial role in various applications. It can be used for confidence assessment, quantifying information gain, and detecting outliers.
Notably, in \gls{nerf} for example, NeRF-W~\cite{Martin2021NeRFW} leverages uncertainty estimation to mitigate the influence of transient scene elements, while S-NeRF~\cite{shen2021snerf} incorporates uncertainty by sampling to encode the posterior distribution across potential radiance fields.
Recent uncertainty-guided methods for view selection~\cite{Ran2022NeurAR, Jin2023NeUNBV, Lee2022UncertaintyNBV, Sunderhauf2022DensityAwareEnsembles, Chen2023SCRNeRF} assume that positioning the camera at the pose with the highest uncertainty will yield the highest reconstruction performance.
Some methods ~\cite{Ran2022NeurAR, Jin2023NeUNBV, Chen2023SCRNeRF} have integrated an uncertainty prediction module within the \gls{nerf} framework.
Considering the cost of incorporating a new module into arbitrary \gls{nerf} model, the other trend of work ~\cite{Lee2022UncertaintyNBV, Sunderhauf2022DensityAwareEnsembles} avoids the modification of existing \gls{nerf} architecture.
They estimate the uncertainty of reconstructed scenes based on predicted density and color.
In contrast, Active-NeRF~\cite{Pan2022ActiveNeRF} considers information gain as the reduction of uncertainty in a candidate view, selecting the candidate view with the most reduction of uncertainty.
However, these uncertainty-based methods directly rely on the quality of estimated uncertainty.

\noindent\textbf{Error-guided methods:}
Rendering errors can be seen as a prior for guiding the ray sampling strategy.
For instance, in the work of Zhang et al.~\cite{Zhang2023FewerRays} rays are directed at pixels with significant color changes and areas with higher rendered color loss, achieving faster \gls{nerf} training without compromising the competitive accuracy.
A distortion-aware scheme~\cite{Otonari2022NonUniformRaySample} is adopted for effectively sampling rays in the 360{\textdegree} scene learning.
In a different context, addressing the problem of \gls{nerf} model conversion via knowledge distillation, PVD-AL~\cite{Fang2023PVD-AL} actively selects views, rays, and points with the largest gap between the student and teacher models to enhance the student's understanding of critical knowledge.
While these efforts may not directly address the question of view selection for the general setting of training a \gls{nerf}, they serve as inspiration for designing a method in an error-guided manner.

\noindent\textbf{Scene coverage-based methods:}
Keyframe selection strategy by maximizing the coverage of the scene is crucial in \gls{slam}~\cite{Zhu2022NICESLAM, Yang2022VoxFusion, Jiang2023H2Mapping} for tackling the forgetting issue. 
For example, NICE-SLAM~\cite{Zhu2022NICESLAM} selects keyframes based on the overlap with existing frames, while H2-Mapping~\cite{Jiang2023H2Mapping} focuses on maximizing the coverage of voxels in the scene.
A progressive camera placement technique~\cite{Koponans2023ProgressiveCamera} is proposed for free-viewpoint navigation.
This technique captures new views to achieve uniformity in ray coverage and angulation within a simulation system.
Nevertheless, these approaches either rely on sensor-captured data like point clouds and depth images or require information within a radiance field.
Differently, we focus on understanding the NeRF rendering performance when trained with different view selection algorithms.

%% file: sec/3_motivation.tex
\section{Motivation -- Providing a Robust Evaluation}
\label{sec:motivation}

\begin{figure}[!t]
    \includegraphics[width=1.03\columnwidth]{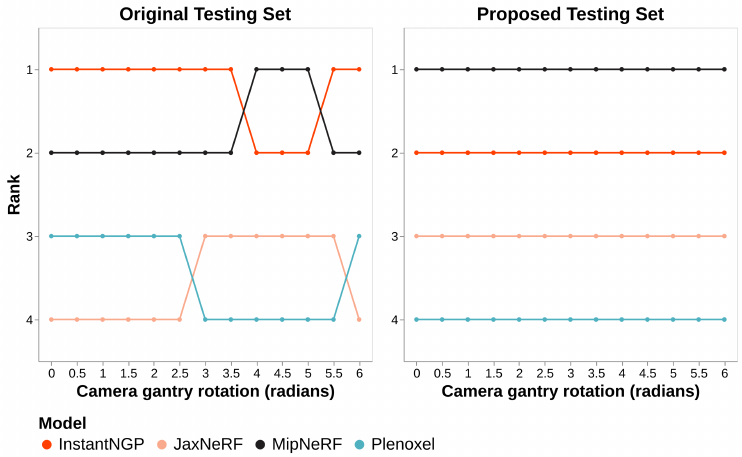}
    \vspace{-15pt}
    \caption{Ranking the rendering performance of four distinct \gls{nerf} models under various \textit{z-axis} rotations of the test camera poses. \textbf{Left}: original test set. \textbf{Right}: proposed test set.}
    \label{fig:rankings}
\vspace{-4pt}
\end{figure}

This section analyzes the importance of view selection whilst evaluating different \gls{nerf} models and illustrates our motivation for investigating an effective view selection algorithm for \gls{nerf}.
To carry out this experiment, we used the widely used NeRF Synthetic dataset~\cite{Mildenhall2020NeRF},
where cameras follow a trajectory resembling a lemniscate in the original test set. Using the Blender files provided by the authors, we generated $12$ additional test sets by rotating the original cameras according to \textit{z-axis}.

We compare the performance of four SOTA \gls{nerf} models---traditional \gls{nerf}~\cite{Mildenhall2020NeRF}, MipNeRF~\cite{Barron2021MipNeRF}, Plenoxels~\cite{Fridovich2022Plenoxels} and InstantNGP~\cite{Muller2022InstantNGP}.
Utilizing the checkpoints provided by the authors or adhering to the same training guidelines, we evaluate their performance across $13$ sets of distinct camera poses.
These models are ranked based on their average \gls{psnr} on each separate test set as visually presented in \Cref{fig:rankings}.
Notably, the ranking exhibited variations across various rotation scenarios; for instance, while InstantNGP excelled as the \gls{sota} approach on the reference test pose, it was outperformed by MipNeRF in certain rotation scenarios. 

To gain a deeper understanding of the impact of camera selection, we propose introducing the coverage density measure,  supported on the mesh $\mathcal{M}$. Given a set of $n$-views $V = \{v_1,\dots,v_n\}$ with associated rays $r^i_{j,k}$ for the $j,k$-th pixel of the $i$-th view, we compute the coverage measure $\mathfrak{C}$ defined by,
\vspace{-6pt}
\begin{align}
    \mathfrak{C}(\mathcal{M},V) &= \frac{1}{\kappa} \sum_{l=1}^M \mathcal{\delta}_{\mathbf{x}_l} \mathcal{W}(\mathcal{M},V,\mathbf{x}_l),&\\
    \mathcal{W}(\mathcal{M},V,\mathbf{x}_l)&=\Big| \Big(\ \sum_{i=1}^n\sum_{\substack{1\le j\le H \\ 1\le k\le W}} (r^i_{j,k} \stackrel{1}{\cap} \mathcal{M})\Big) \cap B_2^\ell(\mathbf{x}_l)\Big| , \nonumber
\end{align}
with $\kappa$ a normalization factor, $(\mathbf{x}_l)_{l\in 1\cdots M}$ a uniform point-cloud discretizing $\mathcal{M}$~\cite{lebrat2021mongenet}, where $B_2^\ell(\mathbf{x})$ is the $L_2$-ball of radius $\ell$ centered in $\mathbf{x}$ and $r\cap^1\mathcal{M}$ denotes the first intersection between the ray $r$ and the mesh $\mathcal{M}$. We display $\mathfrak{C}(\mathcal{M},V_\text{default})$ and $\mathfrak{C}(\mathcal{M},V_{\text{rot}_{90^{\circ}}})$ in \Cref{fig:motivation-a}.

\begin{figure}[t]
    \centering
    \begin{tikzpicture}
        \node[inner sep=0pt] (tikzmagical) at (0,0) 
        {\includegraphics[width=.95\columnwidth]{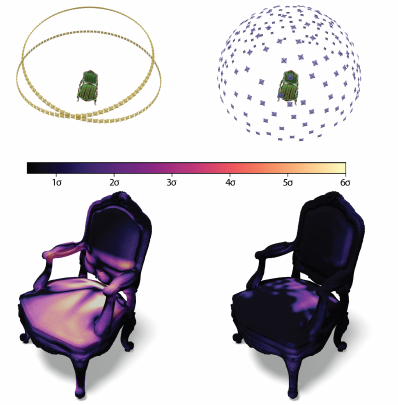}};
        \node[rotate=90, anchor=center,scale=1.00,text=lionY] at (-3.9,2.4) {original};
        \node[rotate=90, anchor=center,scale=1.00,text=lionP] at (0.15,2.4) {proposed};
        \node[rotate=0, anchor=center,scale=.82] at (-0.0,.98) {$|\mathfrak{C}(\mathcal{M},V_{\text{default}}) -\mathfrak{C}(\mathcal{M},V_{\text{rot}_{90^{\circ}}}) | $};
    \end{tikzpicture}
    \caption{Visual comparison between original (left) and proposed (right) test set for \gls{nerf} Synthetic dataset. The top row visualizes the default (w/o. rotation) test cameras' distribution in the 3D space. The bottom displays the absolute difference of the coverage density measure between default and $90^\circ$. A lighter color indicates higher discrepancies in terms of the standard deviation $\sigma$.}
    \label{fig:abs-diff}
    \vspace{-7pt}
\end{figure}

To circumvent the biased evaluation methodology, we devised a new uniform test set where all cameras are evenly distributed on a sphere centered on the reconstructed object and displayed in the~\Cref{fig:abs-diff} (top).
We visualize the absolute difference of $\mathfrak{C}(\mathcal{M},V_\text{default})$ and $\mathfrak{C}(\mathcal{M},V_{\text{rot}_{90^{\circ}}})$ for the original and the proposed test set in~\Cref{fig:abs-diff} (bottom).
We can observe that the proposed test set provides an even coverage of the scene, and thereby, effectively enables a robust evaluation shown in~\Cref{fig:rankings} (right).

The effectiveness of view selection on test sets also motivates us to explore and develop an effective method for view selection on training sets that yields \gls{nerf} with improved rendering performance.
We develop this idea in~\Cref{sec:method}.


%% file: sec/4_method.tex
\section{NeRF Director}
\label{sec:method}
This paper examines strategies to achieve optimal rendering quality by selecting training views from an extensive collection of images. Formally,
given a set of training views $V = \{v_1, v_2, ..., v_N\}$ and a target view budget $n$, our objective is to select a subset of training views $S$ that maximizes the rendering performance $\cal Q$ of \gls{nerf}.
This problem can be formally defined as,
\vspace{-6pt}
\begin{equation}
    \argmax_{S \subseteq V} ~\mathcal{Q} (S),
\end{equation}
where $|S| = n$.
To tackle this problem, this paper proposes two view selection methods --- \acrfull{fvs} and \acrfull{hs}.

\subsection{Farthest View Sampling}
\label{sec:fps}

\Gls{fps}~\cite{Eldar1997FPS} is a classical algorithm typically applied to select a subset of points from extensive raw point cloud data.
In this work, we adapt this algorithm for the view selection problem, aiming to efficiently capture features of the target scene by selecting representative views that are as different from each other as possible.

\Cref{alg:fvs} outlines our proposed \gls{fvs}.
Given a large set of training views $V$ and an initially selected subset of views $S$, our \gls{fvs} first identifies the nearest selected neighbor of each point in the remaining set $V\setminus S$. 
The algorithm then selects the candidate view with the maximum distance as the next addition to the subset $S$.
The employed distance metric considers both the spatial expansion of cameras and the diversity of scene features captured by views.

\begin{algorithm}[!t]
    \DontPrintSemicolon
    \SetKwInOut{Input}{input}\SetKwInOut{Output}{output}
    \SetKwComment{Comment}{/*}{*/}

    \Input{$V \in v^n$ , $n \in \mathbb{N}$, $k \in \mathbb{N}$ \\
    $\mathbf{d}(\bullet ,\bullet ) : v \times v \rightarrow \mathbb{R}^+$
    }
    \Output{$S$}
    \BlankLine
        
    $S \gets \texttt{random\_sampler}(k, V)$ \\
    \While{$|S| < n$}{
        $v^* \gets \displaystyle{\argmax_{v \in V \setminus S}\ \left[\min_{s \in S}\ \mathbf{d}(v, s)\right]}$
        \\
        \BlankLine
        $S \gets S \cup \{v^*\}$
    }
    \textbf{return $S$}
    \caption{Farthest View Sampling.}
    \label{alg:fvs}
\vspace{-4pt}
\end{algorithm}

\vspace{-8pt}
\paragraph{Spatial expansion of cameras:}
We measure the spatial distance across cameras, denoted as $d_{spatial}$, by evaluating the distance between the camera centers of the candidate view $c_{v}$, and the selected view's camera center $c_{s}$.
When all training views are captured from a common sphere, we adopt the great-circle distance $d_{gc}$ as a metric. The distance is defined as,
\vspace{-4pt}
\begin{equation}
    \mathbf{d}_{gc}(c_v, c_s) = \arccos \left( c_v \cdot c_s \right).
\end{equation}

In cases where training cameras are distributed throughout the scene, we employ the  Euclidean distance to measure spatial separation,
\vspace{-4pt}
\begin{equation}
    \mathbf{d}_{euc}(c_v, c_s) = ||c_v - c_s||_2^2.
    \label{equation:d_euc}
\end{equation}

\vspace{-8pt}
\paragraph{Scene diversity depicted in views:}
In an uncontrolled environment, the training images' perspective can be skewed away from the scene's central point, leading nearby cameras to capture entirely different visual content of the scene, e.g. two cameras at the same position orienting in two different directions.
To address this challenge, we leverage the sparse 3D point cloud and 2D image correspondences computed by the underlying \gls{sfm} algorithm used to generate the poses of the training images. 

Let $ {\cal A} \in \mathbb{N}^{N\times N}$ be a symmetric matrix of pair-wise view similarities. We denote the similarity $\mathcal{A}_{ij}$ as the count of 3D points in the SfM's sparse point cloud, triangulated from 2D feature correspondences between views $i$ and $j$. This measure takes into account visual content, field of view, and relative camera positioning between views. Then, using ${\cal A}$, the view photogrammetric distance $\mathbf{d}_{photo}$ is defined as,
\begin{equation}
    \mathbf{d}_{photo}(v_{i}, v_{j}) = 1 - \frac{\mathcal{A}_{ij}}{\max(\mathcal{A})}.
    \label{equation:A}
\end{equation}

Overall, our view distance $\mathbf{d}(\bullet, \bullet)$ is composed of two parts --- $\mathbf{d}_{spatial}$ considering the spatial distance between camera centers of two views (either $\mathbf{d}_{gc}$ or $\mathbf{d}_{euc}$), and $\mathbf{d}_{photo}$ representing the difference in perception content about the scene.
It can be expressed as,
\begin{equation}
    \mathbf{d}(\bullet, \bullet) = \mathbf{d}_{spatial} + \alpha \ \mathbf{d}_{photo},
    \label{equation:d_photo}
\end{equation}
where $\alpha$ is a positive hyper-parameter associated to photogrammetric distance.

\subsection{Information Gain-based Sampling}
\label{sec:heuristic}

While \gls{fvs} selects $n$ views from $V$ based on a metric $\mathbf{d}$ without training a NeRF model, \Gls{hs} is an incremental procedure deriving information gain from a checkpointed model to select novel views. As detailed in \Cref{alg:hs},
\Gls{hs} begins by randomly sampling $k$ views from $V$ to establish the initial set of training views $S$. Subsequently, at each iteration $i$, it trains a \gls{nerf} model on $S$ and evaluates the remaining views in $V\setminus S$. Employing different evaluation measures and sampling algorithms, it augments the current training set $S$ by selecting $l_i$ novel views from $V\setminus S$. Optionally, a density relaxation step can be performed on this augmented training set. This process continues until $n$ new views are selected.

\begin{algorithm}[!t]
    \DontPrintSemicolon
    
    \SetKwInOut{Input}{input}\SetKwInOut{Output}{output}
    \SetKwComment{Comment}{/* }{ */}
    
    \Input{$V \in v^n$, $n \in \mathbb{N}$, $k \in \mathbb{N}$ \\
    $(l_i)_{i=1\dots m}$ \flushright \vspace{-1em}\includegraphics[height=1em]{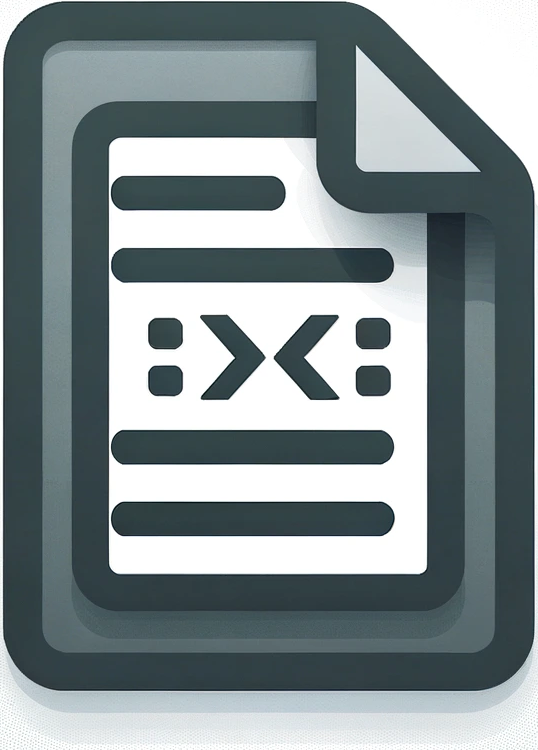} Number of added view list\\
    }
    \Output{$S$}
    \BlankLine
    
    $S \gets \texttt{random\_sampler}(k, V)$
    
    \For{$i \gets 1$ \KwTo $m$} {
        $f_{\theta^\star} \gets \texttt{model\_trainer}(S,f_\theta)$ \;
        $\mu \,\,\,\, \gets  \texttt{model\_evaluator}(f_{\theta^\star}, V \setminus S)$ \;
        $S^* \,\gets \mathit{probabilitySampler}(\mathbb{P}_\mu, V \setminus S, l_i)$ \newline $\phantom{lll\gets}$ \includegraphics[height=1em] {fig/comment_emo.png} Defined in \Cref{sec:vmf}.\; 
        $S^* \,\gets \mathit{relaxation}(S^*, S)$ \newline $\phantom{lll\gets}$ \includegraphics[height=1em] {fig/comment_emo.png} (Optional) see \Cref{sec:lloyd}.\;
        $S \,\,\,\,\gets S \cup S^*$ \;
    }
    \textbf{return} $S$
    \caption{Information Gain-based Sampling.}
    \label{alg:hs}
\vspace{-4pt}
\end{algorithm}

\vspace{-3pt}
\subsubsection{Sampling Views Maximizing Information Gain}
\label{sec:vmf}
At the core of this heuristic-based procedure lie two crucial components: the definition of information gain and the method for sampling views according to this quantity.

\paragraph{Target density construction:}
We propose to measure the error, in terms of the PSNR ranking, for every single remaining training view.

\paragraph{Sampling from the remaining view:}
We first introduce the Zipf view sampler. 
Given a set of $q$ views $\{ v_1, \dots, v_q \}$ with associated error or uncertainty measurements $\mathbf{m}=(m_1,\dots,m_q)$, we first define the greedy selection $k$ as,
\begin{equation}
    S^*  \stackrel{k}{\sim} \mathcal{Z}\text{exp}(\mathbf{m}),
\end{equation}
where $\stackrel{k}{\sim}$ describes the random selection of $k$ elements without replacement, and $\mathcal{Z}\text{exp}$ is a Pareto-Zipf law~\cite{powers1998zipflaw} with exponential weighting. The probability mass function is defined by,
\begin{equation}
f_i \propto \frac{e^{-\gamma \frac{\text{rank}(m_i)}{q-1}}}{K_q},
\end{equation}
with $K_q$ a normalization factor, and $\gamma$ a hyper-parameter controlling the sampling randomness. When $\gamma \rightarrow \infty$, this method is equivalent to the deterministic greedy sampler; conversely, when $\gamma \rightarrow 0$, this method is equivalent to \gls{rs}.

As an alternative approach to exploring possible interactions between nearby error measurements within a probabilistic framework, we introduce the (M-vMF) view sampler. This sampler is built on a categorical mixture of the \gls{vmf} distribution~\cite{banerjee2005clustering}. This sampler assumes that each already sampled view induces a vMF distribution centered at its respective camera position on the unit sphere. The probability density function of this mixture model is given by,
\begin{equation}
 f(x; \mathbf{v}, \mathbf{m}, \kappa) = \sum_{i=1}^{q} \alpha_i g(x; v_i, \kappa),
\end{equation}
where $g(x;v_i, \kappa)$ is the induced vMF probability density function for view $v_i$, with a shared concentration hyper-parameter $\kappa$. The parameter $\kappa$ regulates the dispersion of the distributions around their mode at the camera center of $v_i$. As $\kappa \rightarrow \infty$ the vMF density approaches a delta function located at $v_i$'s camera center; conversely, when $\kappa \rightarrow 0$ this method is equivalent to a uniform distribution over a unit-sphere.
Mathematically,
\begin{equation}
 g(x; v_i, \kappa) = c_3(\kappa) \text{exp}(\kappa v_i^Tx),
\end{equation}
where $c_3(\kappa)$ is the standard vMF normalization term.
The blending of these vMF components is controlled by the weights $\mathbf{\alpha} = (\alpha_1, \dots, \alpha_q)$, obtained through a softmax function applied to error measurements $\mathbf{m}$ with a temperature parameter $\sigma$. Specifically,
\begin{equation}
 \alpha_i = \frac{\text{exp}\left(\frac{\hat{m}_i}{\sigma}\right)}{\sum_{j=1}^q \text{exp}\left(\frac{\hat{m}_j}{\sigma}\right)},
\end{equation}
where $\hat{m}_i=\frac{\max(\mathbf{m}) - m_i}{\max(\mathbf{m}) - \min(\mathbf{m})}$ is an inverse ranking function of the view's error, min-max normalized between zero and one. 

To sample a new view location using this model, we begin by sampling from the categorical distribution controlled by $\mathbf{\alpha}$. Subsequently, we sample a 3D point from the corresponding vMF distribution. As this process does not ensure the existence of a view at the sampled location, we assign the closest view in $V \setminus S$ as the sampled view. Within this framework, regions with views of higher errors are more likely to be sampled. 

\vspace{-1pt}
\subsubsection{Avoiding Oversampling Complex Object Parts}
\label{sec:lloyd}

As further described in \Cref{sec:vmf}, we observe that purely greedy approaches tend to produce clusters of cameras in particular regions of the 3D space. 
Indeed, a scene may comprise more challenging parts to learn by neural-rendering primitive. Uncertainty or error will be more substantial for this specific scenario, resulting in~\Cref{alg:hs}'s proposal of novel training views clustering in this area. 
This over-exploitation behavior is detrimental; we empirically observe that it can lower the performance of the view proposal algorithm below that of the baseline \gls{rs}.
To tackle this problem, we introduce a relaxation step after the proposal of novel views via the view sampler, as described in~\Cref{alg:hs}. We adapt the Lloyd-Max algorithm\cite{lloyd1982lloyd}, commonly used for quantization, to uniformize the placement of the newly proposed camera. More specifically, we propose to build a uniform probability distribution whose support is defined by the convex hull of all available training cameras. After the Voronoi tessellation construction, this distribution is used to compute each cell's centroid. We apply the Lloyd iteration only to the new subset of camera $S^*$. More details on the implementation can be found in Supplementary.

%% file: sec/5_exp.tex
\section{Experiments}
\label{sec:exp}
\input{fig/ingp_overview}
\subsection{Experimental Setup}
\label{sec:setup}

We experimented on two widely used datasets: \gls{nerf} Synthetic~\cite{Mildenhall2020NeRF} and TanksAndTemples~\cite{knapitsch2017tnt}.

\paragraph{NeRF Synthetic:}
It contains $5$ synthetic objects.\footnote{The dataset originally contains eight scenes, but with the data provided, we only managed to reproduce the original rendering quality for five of them.}
We reproduced the exact rendering settings and kept the original image resolution proposed in~\cite{Mildenhall2020NeRF}.
Each scene comprises $200$ test images sampled as described in~\Cref{sec:motivation} and a pool of $300$ views evenly distributed for training.
We generated ten training sets to ensure reproducibility and statistical significance. 

\paragraph{TanksAndTemples:}
It is a real-world dataset containing $4$ scenes.
Each scene comprises $251$ to $313$ training images and $25$ to $43$ test images.
Due to significant bias in the testing view (see. Supplementary),
we opted to combine original training and test images and resplit them while keeping the same number of test images for each scene.
We followed the method described in \Cref{alg:fvs} to sample the new test set and kept the rest of the views for training.

\paragraph{Backbones and evaluation metrics:}
We conducted our experiment and analysis based on two SOTA \gls{nerf} models --- InstantNGP~\cite{Muller2022InstantNGP} and Plenoxels~\cite{Fridovich2022Plenoxels}.
As evaluation metrics, we consider \acrfull{psnr} ({\color{red}{$\uparrow$}}) and \acrfull{ssim} ({\color{red}{$\uparrow$}})~\cite{wang2004ssim}.

\subsection{Implementation Details}
\label{sec:detail}
We conduct a series of experiments with five repetitions using different random seeds and varying training/test sets for synthetic scenes.
The process begins by randomly selecting an initial set of $5$ views.
Subsequently, we add $5$ more views in each step, reaching a total of $30$ views.
The view selection process continues by adding $10$ more views at each step until accumulating a total of $150$ views.
We train each model from scratch for each view selection choice and report novel-view rendering performance on our test set.
We report results for \gls{rs} and our proposed \gls{fvs} and \gls{hs}, and for a comprehensive benchmark, we implement and compare two uncertainty-based \gls{hs} variants --- ActiveNeRF~\cite{Pan2022ActiveNeRF} and Density-aware NeRF Ensembles~\cite{Sunderhauf2022DensityAwareEnsembles}.
These variants are built upon the InstantNGP backbone, and their implementation details are provided in Supplementary.

\subsection{Evaluation on NeRF Synthetic Dataset}
\label{sec:exp/blender}

The results in terms of PSNR and SSIM for the InstantNGP backbone and an increasing number of training views across different view selection methods are depicted in \Cref{fig:blender_psnr} and \Cref{fig:blender_ssim}.
It can be observed that \gls{fvs} and \gls{hs} significantly outperform other view selection methods. Conversely, view selection methods from ActiveNeRF and Density-aware NeRF Ensembles exhibit inferior performance compared to \gls{rs}.
We attribute this gap in performance to two main factors: first, the uncertainty predicted does not consistently correlate with the expected reconstruction improvement for the candidate view; second, in certain scene areas, adding more views does not always result in decreased uncertainty, leading to oversampling, which is not rectified by spatial regularization.
Similar results are obtained for the Plenoxel backbone and are provided in Supplementary.
We also provide the runtime cost analysis in Supplementary, showing that our proposed \gls{fvs} can reach converged quality more efficiently than \gls{rs} under the same view budget.

\subsection{Evaluation on TankAndTemples Dataset}
\label{sec:exp/tnt}

We extended our experiments to the TanksAndTemples dataset to assess the impact of different view selection methods on rendering performance for real-world data.
\Cref{fig:tnt_psnr} and \Cref{fig:tnt_ssim} display the experimental results in terms of PSNR and SSIM.
Notably, when the training view budget exceeds $30$ views, \gls{fvs} demonstrates superior performance, followed by \gls{hs}.
In contrast to the results with the NeRF Synthetic dataset, view selection based on Density-aware NeRF Ensembles achieves better performance than \gls{rs} in a higher view number regime (more than 60 views).
This could be attributed to the improved uncertainty quantification on realistic data of the Density-aware NeRF Ensembles, where candidate views are not uniformly distributed.
Intriguingly, ActiveNeRF provides the lowest performances for our test settings, and we attribute this to its exploration of a distinct training regime for NeRF (less than 30 views) and the limited pool of training views considered (100 views) as indicated in a recent study~\cite{Koponans2023ProgressiveCamera}.

\input{fig/ablation}

\subsection{Ablation Study}
\label{sec:exp/abalation}

This section introduces our ablation experiments on TanksAndTemples dataset to make a comprehensive analysis of the design of our proposed \gls{fvs} and \gls{hs}.
We use InstantNGP~\cite{Muller2022InstantNGP} as our backbone and PSNR as the evaluation metric. 
We conducted experiments following the description in~\Cref{sec:detail}.

\subsubsection{Information Gain-based Sampling}
\label{sec:ablation/info}

\paragraph{Information type:}
There are two potential information types for \gls{hs} methods: error and uncertainty.
We first explore the impact of different information gains on the performance of \gls{hs}.
We implemented a variant of \gls{hs} based on uncertainty, which was quantified through Density-aware NeRF Ensembles~\cite{Sunderhauf2022DensityAwareEnsembles}. 
Both error and uncertainty variants utilized the \gls{vmf} view sampler with applied relaxation.
\Cref{fig:tnt_info} provides a quantitative visualization of the comparison results.
Error-based \gls{hs} consistently outperforms uncertainty-based \gls{hs}. 

\begin{figure}[!t]
    \centering
    \includegraphics[width=0.89\columnwidth]{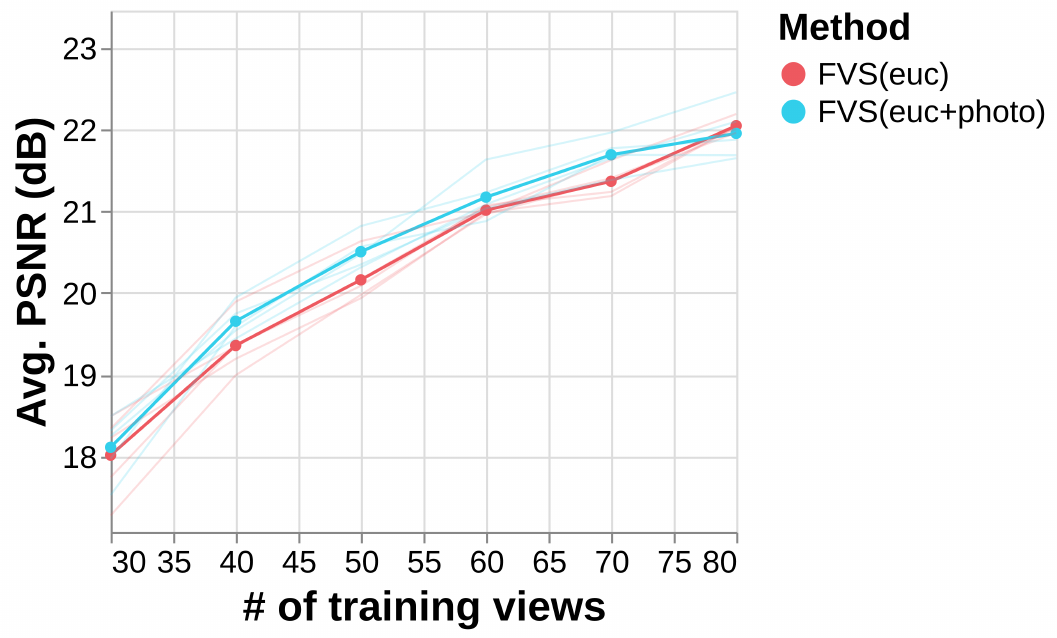}
    \caption{Comparison results between \gls{fvs} ($\mathbf{d}_{euc}$) and \gls{fvs} ($\mathbf{d}_{euc} + \mathbf{d}_{photo}$) in terms of PSNR on the scene \textit{Playground}.}
    \label{fig:playground}
\vspace{-15pt}
\end{figure}

\paragraph{Sampling strategy:}
We further investigate the impact of different probabilistic mass functions on the \gls{hs}.
We compared the M-\gls{vmf} and the Zipf view samplers.
Two Zipf view samplers were implemented with different $\gamma$ settings: $\gamma = 10$ and $\gamma \rightarrow \infty$, representing the deterministic greedy. 
Quantitative results shown in \Cref{fig:tnt_hs} indicate that view samplers, when combined with relaxation and consideration of nearby error measurements, can effectively select training views, thereby enhancing the performance of a \gls{nerf}.

\subsubsection{Farthest View Sampling}
\label{sec:ablation/fvs}

We explore four combinations of spatial distance $\mathbf{d}_{spatial}$ and photogrammetric distance $\mathbf{d}_{photo}$.
Specifically, we compare only $\mathbf{d}_{spatial}$ based on the great circle and the Euclidean distance, $\mathbf{d}_{gc}$ and $\mathbf{d}_{euc}$ respectively, as well as these two spatial distances separately combined with $\mathbf{d}_{photo}$.
For methods using $\mathbf{d}_{gcd}$, we project all training views' camera centers onto a common sphere.
The average quantitative results are presented in~\Cref{fig:tnt_dist}. 
It is evident that selecting views solely based on $\mathbf{d}_{gcd}$ could be insufficient for complex real-world datasets. 

\paragraph{Discussion on the use of $\mathbf{d}_{photo}$:}
When looking at the results of each scene, we notice an interesting case in \textit{Playground} shown in~\Cref{fig:playground}, where cameras are distributed across the space, yet most share similar attention regions of the scene.

In such cases, the information provided by the camera center fails to indicate the area of the scene observed. 
Adopting $\mathbf{d}_{photo}$ is crucial for improving the scene diversity within selected training views. Despite generally offering better performances for our datasets, the use of \gls{fvs} may be limited in complex indoor environments where relying solely on distances between camera centers may not adequately capture view similarity, especially in scenarios involving occlusions like indoor exploration.


%% file: fig/ingp_overview.tex
\begin{figure*}[t]
    \centering
    \begin{minipage}{0.8506\textwidth}
        \centering
        \begin{subfigure}[]{0.49\textwidth}
            \centering
            \includegraphics[width=0.99\textwidth]{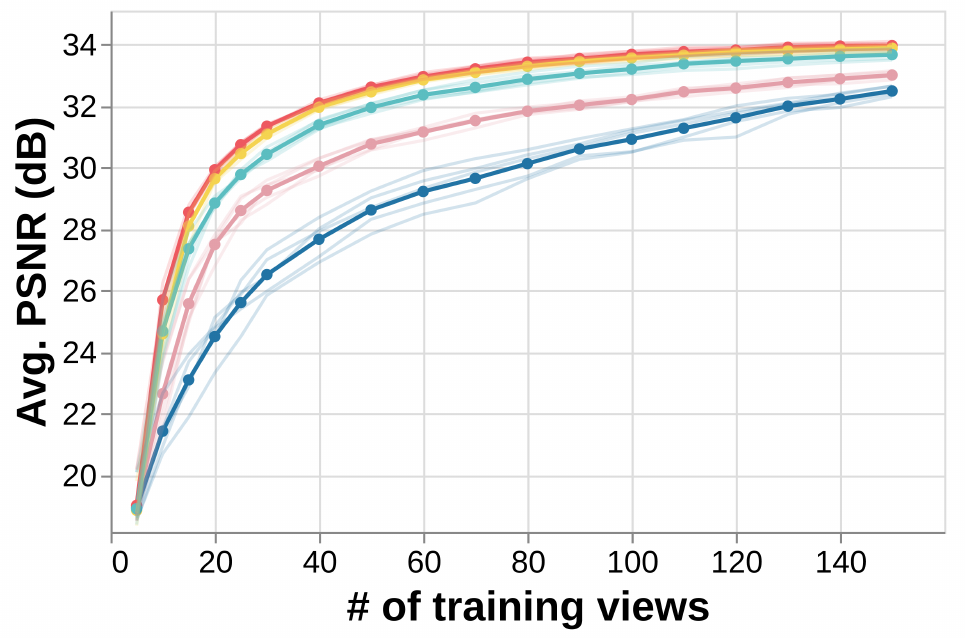}
            \vspace{-17pt}
            \caption{}
            \label{fig:blender_psnr}
        \end{subfigure}
        \hfill
        \begin{subfigure}[]{0.49\textwidth}
            \centering
            \includegraphics[width=0.99\textwidth]{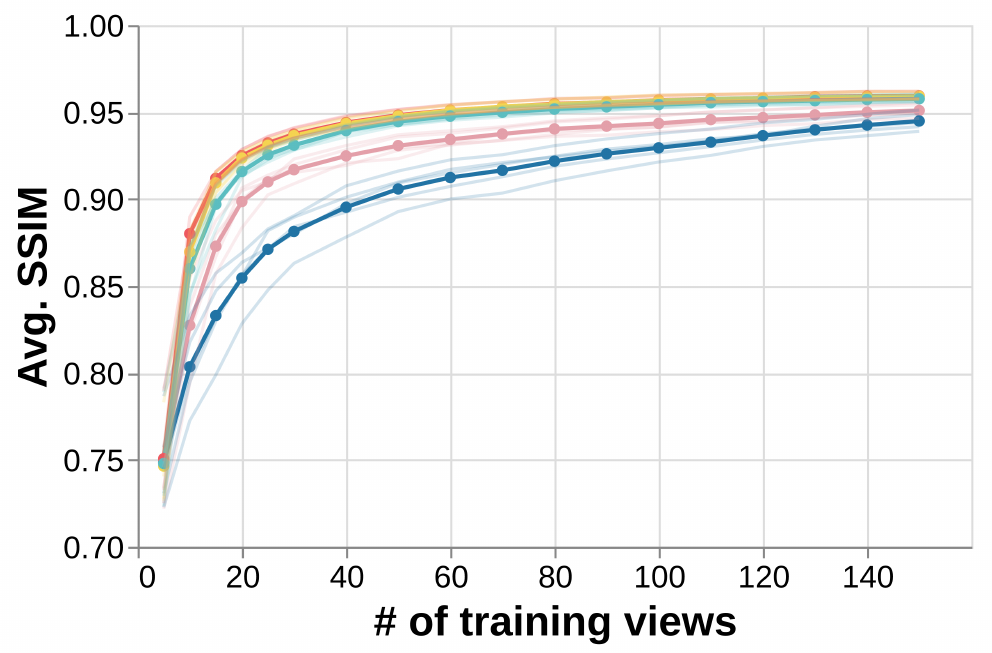}
            \vspace{-17pt}
            \caption{}
            \label{fig:blender_ssim}
        \end{subfigure}
        \hfill
        \begin{subfigure}[]{0.49\textwidth}
            \centering
            \includegraphics[width=0.99\textwidth]{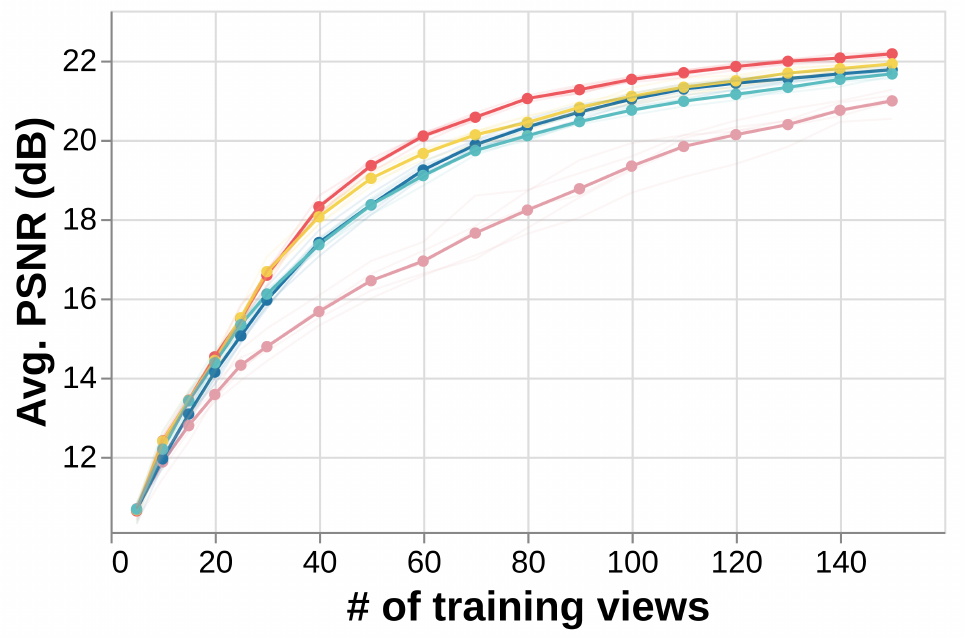}
            \vspace{-17pt}
            \caption{}
            \label{fig:tnt_psnr}
        \end{subfigure}
        \hfill
        \begin{subfigure}[]{0.49\textwidth}
            \centering
            \includegraphics[width=0.99\textwidth]{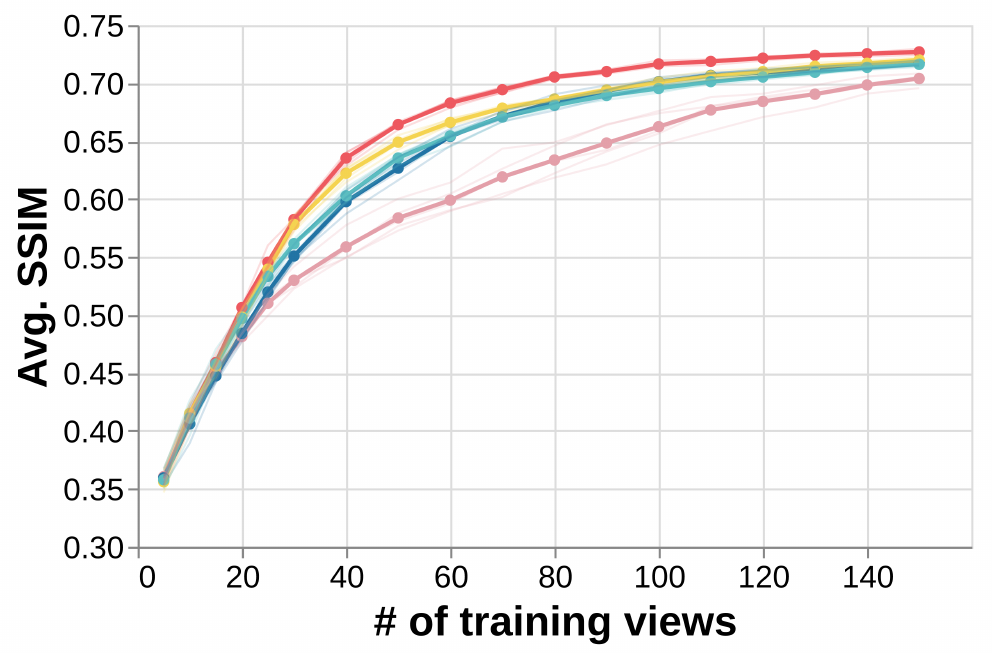}
            \vspace{-17pt}
            \caption{}
            \label{fig:tnt_ssim}
        \end{subfigure}         
    \end{minipage}
    \hfill
    \begin{subfigure}[]{0.136\textwidth}
        \includegraphics[width=1.0\textwidth]{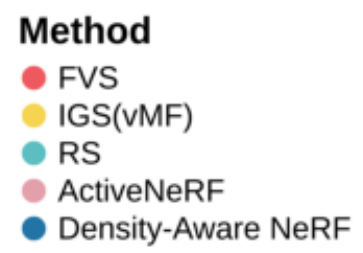}
    \end{subfigure}      
    \vspace{-8pt}
    \caption{Quantitative comparisons of rendering quality along with the increase of used training views sampled by different view selection methods. \textbf{Top}: results on the \gls{nerf} Synthetic dataset in terms of PSNR (a) and SSIM (b). \textbf{Bottom}: results on the TanksAndTemples dataset in terms of PSNR (c) and SSIM(d). Low-opacity lines present the results for each repetition, while high-opacity lines present the average result across five repetitions.}
    \label{fig:overview_results}
\vspace{-10pt}
\end{figure*}

%% file: fig/ablation.tex
\begin{figure*}[t]
    \begin{subfigure}[]{0.31\textwidth}
        \centering
        \includegraphics[width=1.03\textwidth]{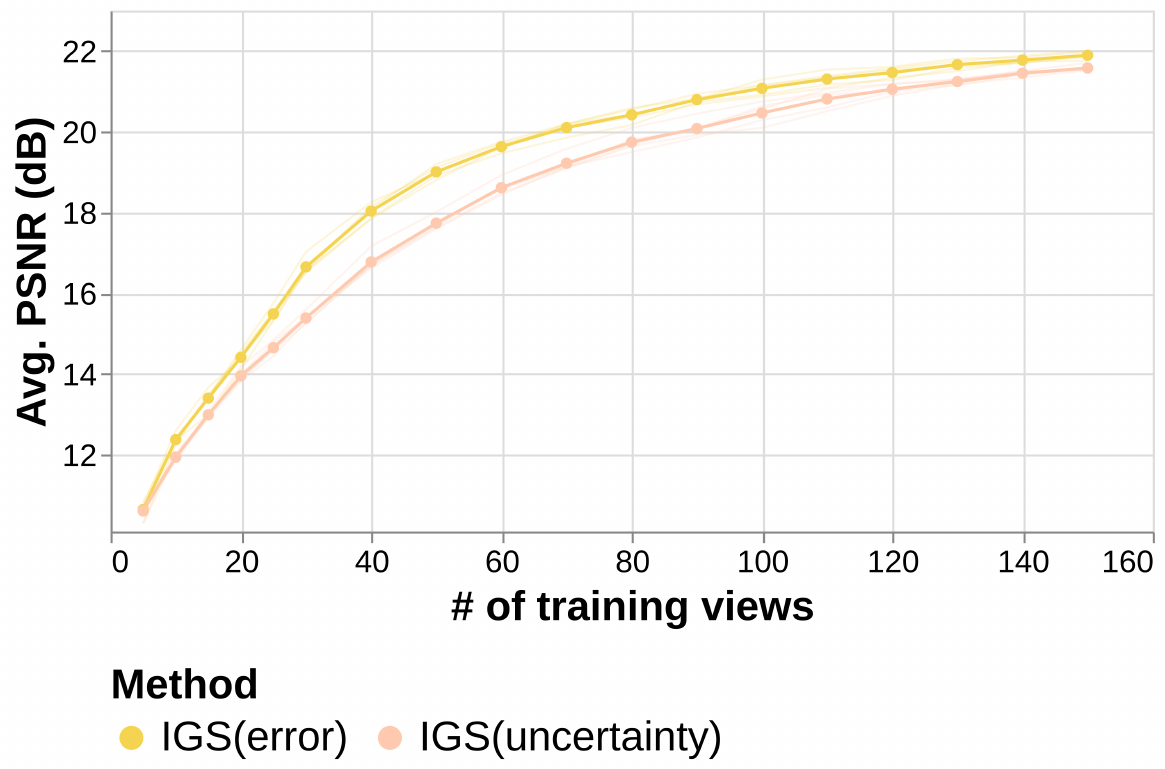}
        \caption{}
        \label{fig:tnt_info}
    \end{subfigure}
    \hfill
    \begin{subfigure}[]{0.31\textwidth}
        \centering
        \includegraphics[width=1.03\textwidth]{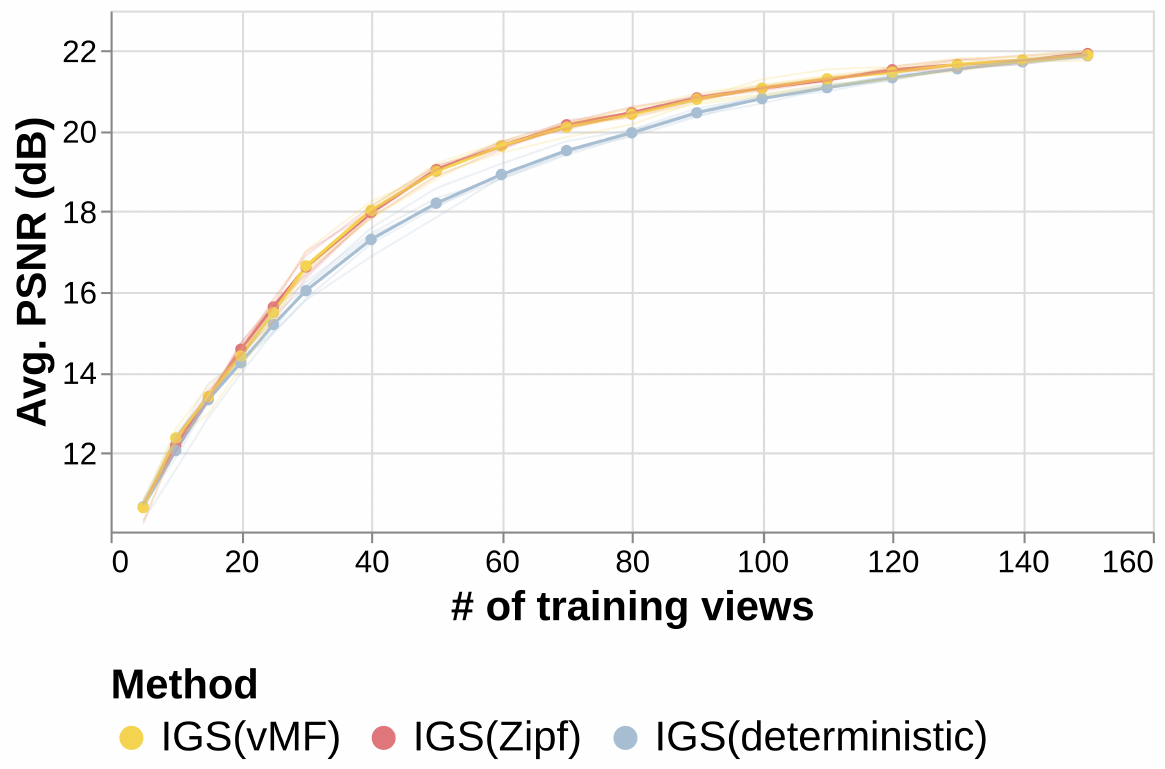}
        \caption{}
        \label{fig:tnt_hs}
    \end{subfigure}
    \hfill
    \begin{subfigure}[]{0.35\textwidth}
        \centering
        \includegraphics[width=1.03\textwidth]{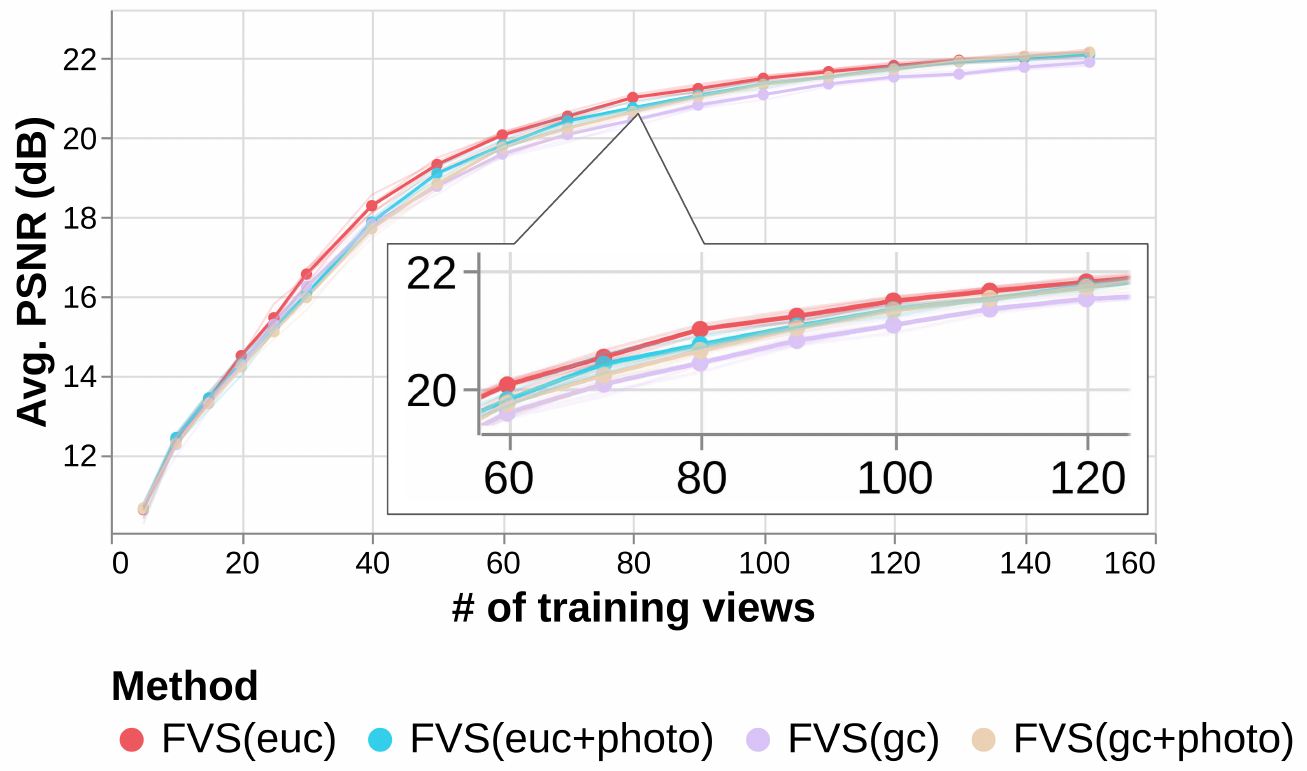}
        \caption{}
        \label{fig:tnt_dist}
    \end{subfigure}    
  
    \caption{Ablation studies of the information type (a) and the sampling strategy (b) in \gls{hs}, as well as different distance metrics in \gls{fvs} (c) on the TanksAndTemple dataset.}
    \label{fig:overview_results}
\vspace{-10pt}
\end{figure*}

%% file: sec/6_summary.tex
\section{Conclusion}
\label{sec:conclusion}

We studied the role of view selection for \gls{nerf} in both training and testing.
We first proposed a novel method to select test views reaching a more robust and reliable evaluation.
We further proposed a novel view selection assessment framework, NeRF Director.
We explored and introduced two view selection methods: \acrfull{fvs}, considering the distance across cameras and the diversity of their content, and an improved \acrfull{hs} approach by incorporating relaxation to avoid clustering.
Our experiments and analysis highlight the role of diversity in selected training views, caution against reliance on information gain-based methods with noisy information, and advocate for spatial relaxation to address sensitivity and cluster-related challenges in information gain-based methods for effective \gls{nerf} learning.
We hope this will serve as a stepping stone in furthering research on this important topic.

\vspace{-7.5pt}
\paragraph{Limitation and Future Works:}
Our proposed methods assume views captured by cameras with the same resolution and are designed under the object-centric setting.
In our future work, we will consider unstructured configurations~\cite{buehler2001UnstructuredLumigraph}.
Also, the quality of available training views is another potential factor impacting \gls{nerf}'s rendering performance.
We will investigate its effect in our future work.

\vspace{-7.5pt}
\paragraph{Acknowledgements:}
The work reported in this paper was partially supported by Australian Research Council (ARC) Discovery grant DP200101942.

%% file: sec/X_suppl.tex
\clearpage
\setcounter{page}{1}
\renewcommand*{\thesection}{\Alph{section}}
\setcounter{section}{0}

 \twocolumn[{%
 \renewcommand\twocolumn[1][]{#1}%
 \maketitlesupplementary
     \centering
     \captionsetup{type=figure}
     \includegraphics[width=0.98\linewidth]{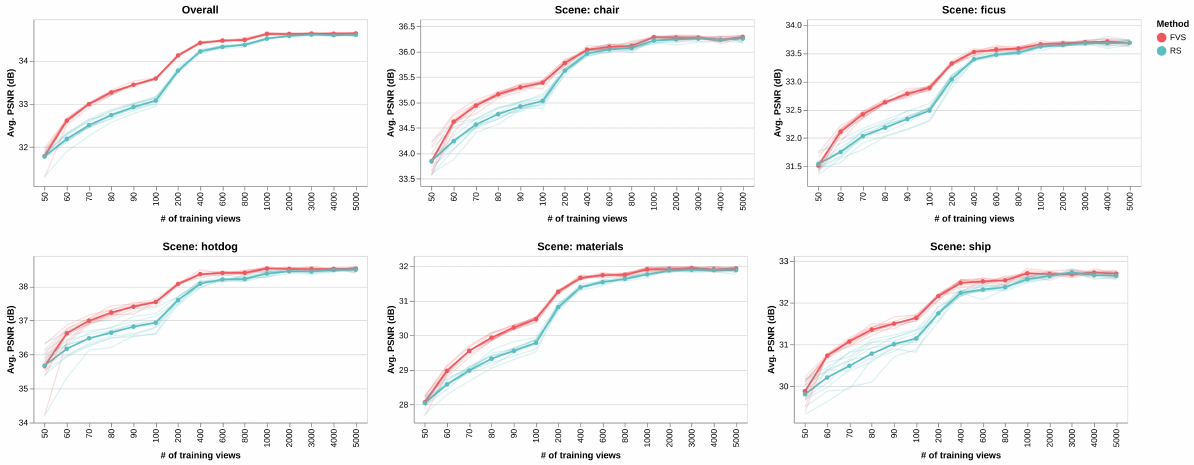}
     \caption{Asymptotic quantitative results of adding more views on NeRF Synthetic datasets in terms of PSNR. The first plot is the overall results across five scenes; the others are scene-specific results. Low-opacity lines present the results for each repetition, while high-opacity lines present the average result across five repetitions.}
     \label{fig:longtest_scenes}
\vspace{8pt}
 }]

\noindent In this supplementary material, we provide details about the following topics:
\begin{itemize}
    \item \textit{Additional details on our motivation} in Appendix~\ref{supp:motivation};
    \item \textit{Relaxation in information gain-based sampling} in Appendix~\ref{supp:lloyd};
    \item \textit{Additional implementation details} in Appendix~\ref{supp:implementation};
    \item \textit{Additional results and visualization} in Appendix~\ref{supp:results};
\end{itemize}

\section{Additional Details on Our Motivation}
\label{supp:motivation}

\paragraph{Asymptotic performance of adding more views.}

To gain a deeper insight into the impact of view selection on novel view synthesis, we train InstantNGP~\cite{Muller2022InstantNGP} on the \gls{nerf} Synthetic dataset with training splits of varied sizes ranging from 50 to 5000 views.  

~\Cref{fig:longtest_scenes} demonstrates overall and scene-specific results. It illustrates that with sufficient training time and sampled views, \gls{rs} achieves the same asymptotic performance as \gls{fvs}. However, it can be noted that \gls{rs} samples cameras independently, which may require more views to achieve the same rendering quality. 

\paragraph{Sparse 3D-reconstruction runtime analysis.}
The proposed technique also offers the advantage of reducing the computation required for the initial sparse reconstruction needed to estimate the camera parameters. Before training any \gls{nerf}, one has to compute camera intrinsics and extrinsics, by solving a \acrlong{sfm} problem, which may become costly as the number of camera $n$ increases. Traditional approaches rely on four steps: feature extraction ($\mathcal{O}(n)$), feature matching ($\mathcal{O}(n^2)$), \gls{sfm} ($\mathcal{O}(n^3)$) and bundle adjustment ($\mathcal{O}(P^3)$), preventing its use for a large number of images.\footnote{where $P$ denotes the number of camera parameters and 3D points.} 

It is worth observing that our current framework and \gls{fvs} could be amenable to performing view selection before solving \gls{sfm}, as the presented algorithm does not require high localization accuracy. For instance, one can imagine a scenario where a real-time slam algorithm (inertial + visual odometry)~\cite{mur2015orbslam, wudenka2021dlrrm, cuVSLAM} estimates the camera poses. Similarly, with the coarse camera overlap, one could swiftly compute the matrix $\mathcal{A}$ adopted in \Cref{equation:A} using recent fast feature matchers~\cite{linden2023lightglue}.

\begin{figure}[h]
    \centering
    \includegraphics[width=0.996\linewidth]{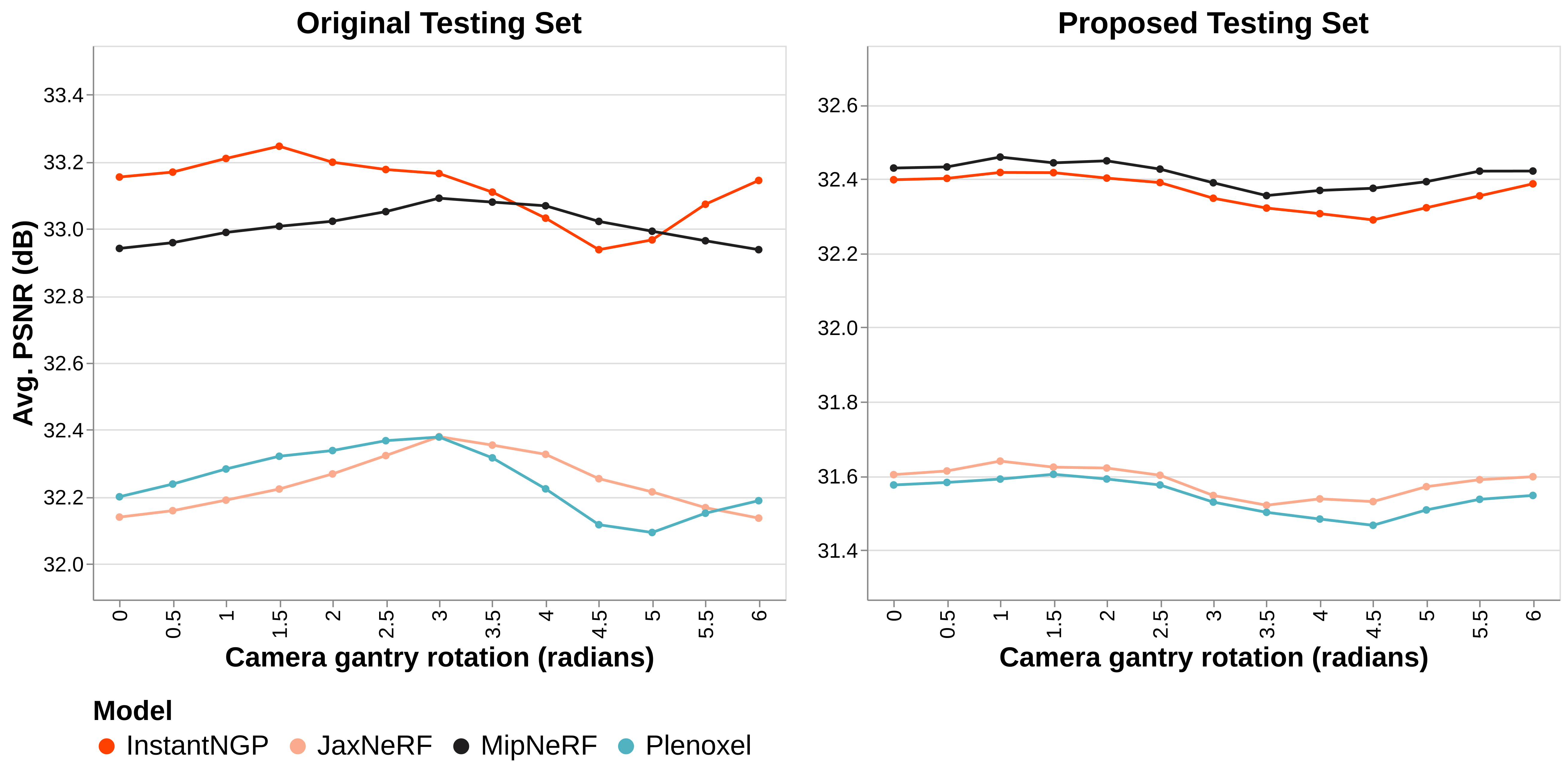}
    \caption{The rendering performance of four distinct \gls{nerf} models, in terms of PSNR, under various \textit{z-axis} rotations of the test camera poses. Left: original test set, Right: proposed test set.}
    \label{fig:ranking_psnr}
    \vspace{-4pt}
\end{figure}

\paragraph{Rankings inversion of SOTA methods.}

As discussed in \Cref{sec:motivation}, view selection is important in the robust evaluation of different \gls{nerf} models. 
~\Cref{fig:ranking_psnr} and \Cref{tab:ranking_psnr} provide detailed quantitative results, in terms of \gls{psnr}, on the original and our proposed test set, each with thirteen sets of rotations.

\begin{table*}[h]\centering
\caption{Quantitative results in terms of PSNR of four SOTA NeRF models under various \textit{z-axis} rotations of the test camera poses.}\label{tab:ranking_psnr}
\scriptsize
\begin{subtable}[h]{\textwidth}
\centering
\begin{tabular}{l|rrrrrrrrrrrrrr}
Original test set &\textbf{0.0} &\textbf{0.5} &\textbf{1.0} &\textbf{1.5} &\textbf{2.0} &\textbf{2.5} &\textbf{3.0} &\textbf{3.5} &\textbf{4.0} &\textbf{4.5} &\textbf{5.0} &\textbf{5.5} &\textbf{6.0} \\\midrule
InstantNGP~\cite{Muller2022InstantNGP} &\cellcolor[HTML]{ea9999}33.15 &\cellcolor[HTML]{ea9999}33.17 &\cellcolor[HTML]{ea9999}33.21 &\cellcolor[HTML]{ea9999}33.25 &\cellcolor[HTML]{ea9999}33.20 &\cellcolor[HTML]{ea9999}33.18 &\cellcolor[HTML]{ea9999}33.16 &\cellcolor[HTML]{ea9999}33.11 &\cellcolor[HTML]{f4cccc}33.03 &\cellcolor[HTML]{f4cccc}32.94 &\cellcolor[HTML]{f4cccc}32.97 &\cellcolor[HTML]{ea9999}33.07 &\cellcolor[HTML]{ea9999}33.14 \\
MipNeRF~\cite{Barron2021MipNeRF} &\cellcolor[HTML]{f4cccc}32.94 &\cellcolor[HTML]{f4cccc}32.96 &\cellcolor[HTML]{f4cccc}32.99 &\cellcolor[HTML]{f4cccc}33.01 &\cellcolor[HTML]{f4cccc}33.02 &\cellcolor[HTML]{f4cccc}33.05 &\cellcolor[HTML]{f4cccc}33.09 &\cellcolor[HTML]{f4cccc}33.08 &\cellcolor[HTML]{ea9999}33.07 &\cellcolor[HTML]{ea9999}33.02 &\cellcolor[HTML]{ea9999}32.99 &\cellcolor[HTML]{f4cccc}32.96 &\cellcolor[HTML]{f4cccc}32.94 \\
JaxNeRF~\cite{Mildenhall2020NeRF} &32.14 &32.16 &32.19 &32.22 &32.27 &32.32 &\cellcolor[HTML]{fff2cc}32.38 &\cellcolor[HTML]{fff2cc}32.35 &\cellcolor[HTML]{fff2cc}32.33 &\cellcolor[HTML]{fff2cc}32.25 &\cellcolor[HTML]{fff2cc}32.21 &\cellcolor[HTML]{fff2cc}32.17 &32.14 \\
Plenoxels~\cite{Fridovich2022Plenoxels} &\cellcolor[HTML]{fff2cc}32.20 &\cellcolor[HTML]{fff2cc}32.24 &\cellcolor[HTML]{fff2cc}32.28 &\cellcolor[HTML]{fff2cc}32.32 &\cellcolor[HTML]{fff2cc}32.34 &\cellcolor[HTML]{fff2cc}32.37 &32.38 &32.32 &32.22 &32.12 &32.09 &32.15 &\cellcolor[HTML]{fff2cc}32.19 \\ 
\end{tabular}
\end{subtable}
\vfill
\vspace{11pt}
\begin{subtable}[h]{\textwidth}
\centering
\begin{tabular}{l|rrrrrrrrrrrrrr}  
Proposed test set &\textbf{0.0} &\textbf{0.5} &\textbf{1.0} &\textbf{1.5} &\textbf{2.0} &\textbf{2.5} &\textbf{3.0} &\textbf{3.5} &\textbf{4.0} &\textbf{4.5} &\textbf{5.0} &\textbf{5.5} &\textbf{6.0} \\\midrule
InstantNGP~\cite{Muller2022InstantNGP} &\cellcolor[HTML]{f4cccc}32.40 &\cellcolor[HTML]{f4cccc}32.40 &\cellcolor[HTML]{f4cccc}32.42 &\cellcolor[HTML]{f4cccc}32.42 &\cellcolor[HTML]{f4cccc}32.40 &\cellcolor[HTML]{f4cccc}32.39 &\cellcolor[HTML]{f4cccc}32.35 &\cellcolor[HTML]{f4cccc}32.32 &\cellcolor[HTML]{f4cccc}32.31 &\cellcolor[HTML]{f4cccc}32.29 &\cellcolor[HTML]{f4cccc}32.32 &\cellcolor[HTML]{f4cccc}32.35 &\cellcolor[HTML]{f4cccc}32.39 \\
MipNeRF~\cite{Barron2021MipNeRF} &\cellcolor[HTML]{ea9999}32.43 &\cellcolor[HTML]{ea9999}32.43 &\cellcolor[HTML]{ea9999}32.46 &\cellcolor[HTML]{ea9999}32.44 &\cellcolor[HTML]{ea9999}32.45 &\cellcolor[HTML]{ea9999}32.43 &\cellcolor[HTML]{ea9999}32.39 &\cellcolor[HTML]{ea9999}32.36 &\cellcolor[HTML]{ea9999}32.37 &\cellcolor[HTML]{ea9999}32.37 &\cellcolor[HTML]{ea9999}32.39 &\cellcolor[HTML]{ea9999}32.42 &\cellcolor[HTML]{ea9999}32.42 \\
JaxNeRF~\cite{Mildenhall2020NeRF} &31.60 &31.61 &31.64 &31.62 &31.62 &31.60 &31.55 &31.52 &31.54 &31.53 &31.57 &31.59 &31.60 \\
Plenoxels~\cite{Fridovich2022Plenoxels} &\cellcolor[HTML]{fff2cc}31.58 &\cellcolor[HTML]{fff2cc}31.58 &\cellcolor[HTML]{fff2cc}31.59 &\cellcolor[HTML]{fff2cc}31.60 &\cellcolor[HTML]{fff2cc}31.59 &\cellcolor[HTML]{fff2cc}31.58 &\cellcolor[HTML]{fff2cc}31.53 &\cellcolor[HTML]{fff2cc}31.50 &\cellcolor[HTML]{fff2cc}31.48 &\cellcolor[HTML]{fff2cc}31.47 &\cellcolor[HTML]{fff2cc}31.51 &\cellcolor[HTML]{fff2cc}31.54 &\cellcolor[HTML]{fff2cc}31.55 \\
\end{tabular}
\end{subtable}
\end{table*}

\section{Relaxation in Information Gain-based Sampling}
\label{supp:lloyd}

\begin{figure}[!h]
    \centering
    \includegraphics[width=0.966\columnwidth]{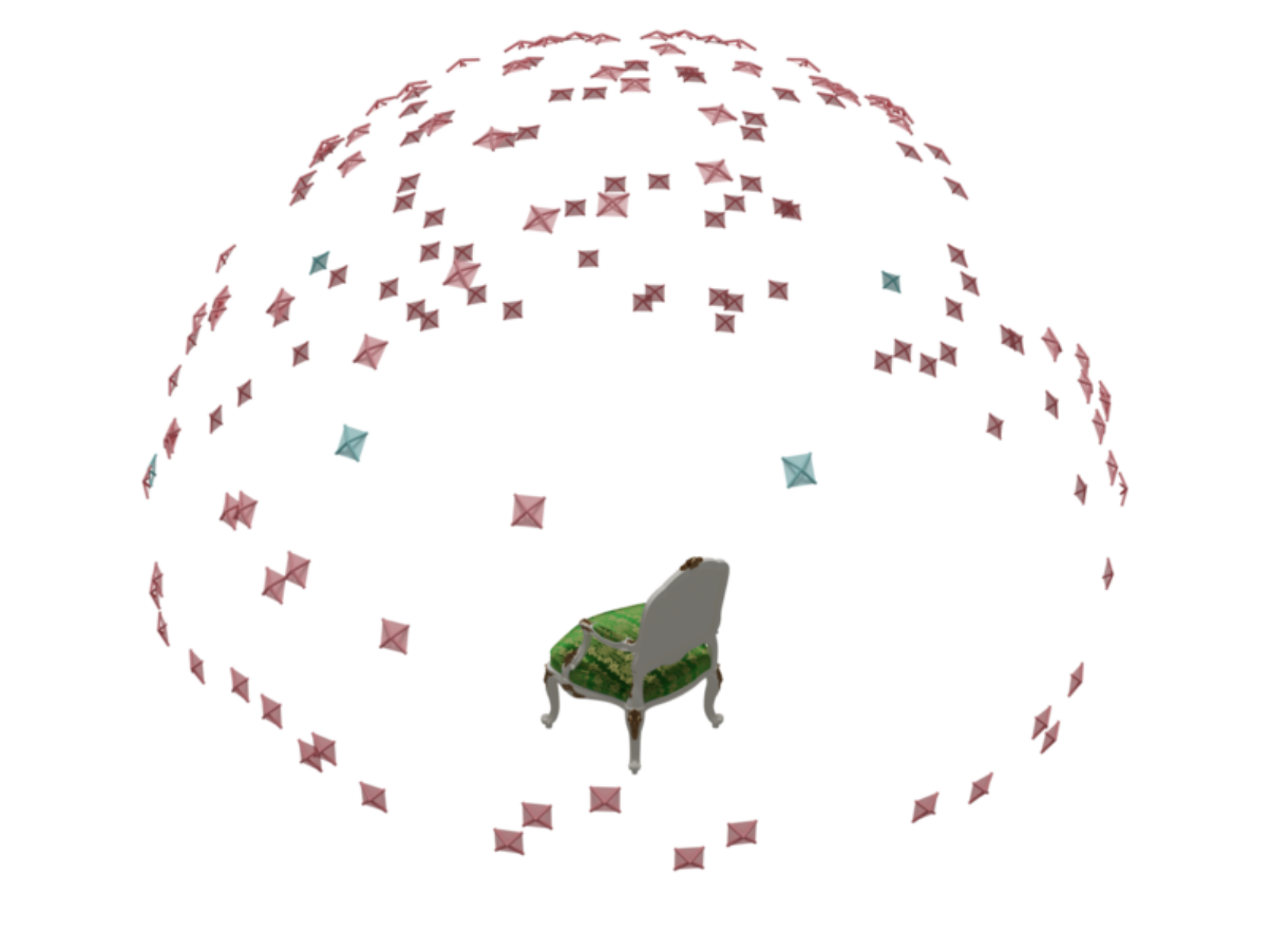}
    \caption{Visualization of the distribution of initial training cameras (in green) and selected cameras (in red) through \gls{hs} without relaxation.}
    \label{fig:oversampling}
    \vspace{-4pt}
\end{figure}

\begin{figure}
    \centering
    \includegraphics[width=0.996\columnwidth]{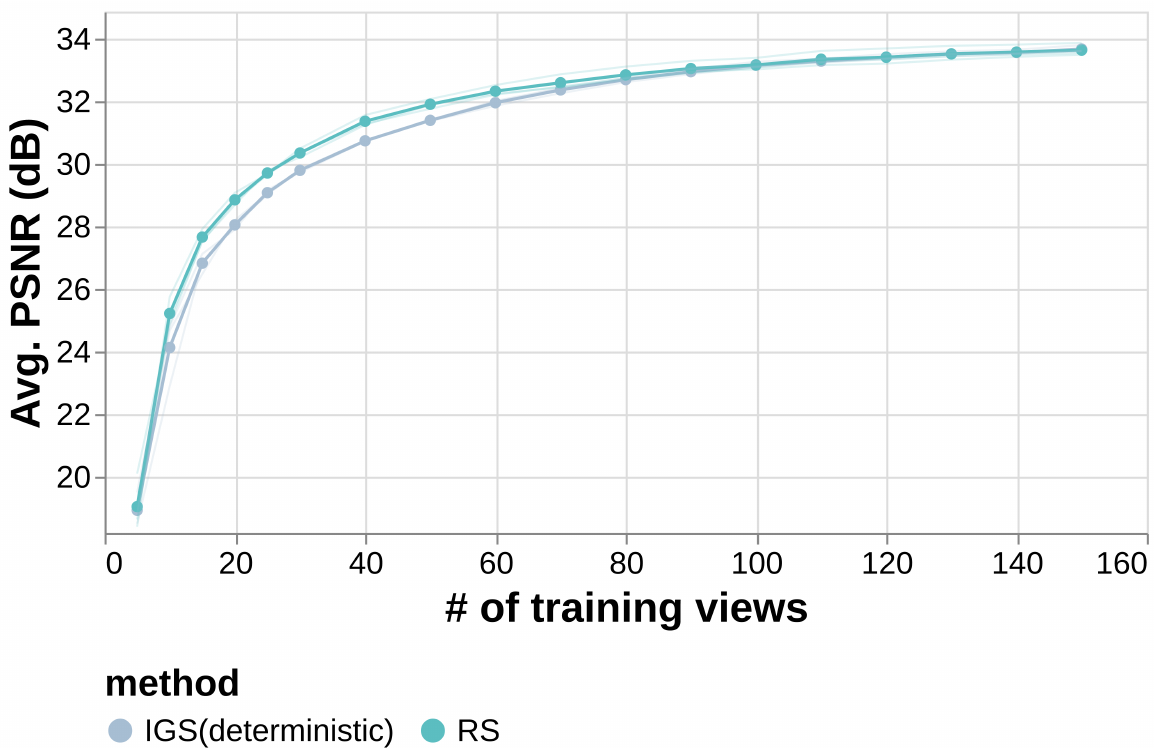}
    \caption{Quantitative comparisons of deterministic \gls{hs} and \gls{rs} on the \gls{nerf} Synthetic dataset. \Gls{hs} without relaxation behaves worse than \gls{rs}.}
    \label{fig:lloyd_supp}
\end{figure}

\paragraph{Limitations without relaxation.}
Varying material or geometry complexity may lead to diverse reconstruction outcomes. \Cref{fig:supp_lloyd} illustrates an example from the \gls{nerf} Synthetic dataset. In this scene, spherical objects in~\Cref{fig:supp_lloyd_GT} have different material complexity. Specifically, intricate surface parts or highly complex materials may contribute to an increased reconstruction error, as shown in~\Cref{fig:supp_lloyd_error}.
As a result, deterministic \gls{hs} methods selecting the view with the highest error or uncertainty tend to stack new training views on these complex areas. For example in~\Cref{fig:oversampling}, newly sampled training views cluster in the forward face of the \textit{chair} due to its increased texture complexity.
%
This overfitting is counterproductive as the performances are inherently inferior.
This can be seen in~\Cref{fig:lloyd_supp} where deterministic \gls{hs} exhibits worse performance than \gls{rs}.
\input{fig/supp_lloyd_motivation}

\paragraph{Lloyd relaxation.}
To alleviate the aforementioned over-sampling effect, we propose a modification based on the LLoyd-Max Algorithm~\cite{lloyd1982lloyd} inspired by optimal transport and stippling theory~\cite{de2012bluenoise, peyre2019optimaltransport}.

More specifically, given a set of $k$ selected views with camera centers $(\mathbf{c}_1,\dots,\mathbf{c}_k)$ and $m$ proposed views, we offer a modification of the Lloyd iteration described in~\Cref{alg:lloyd}.

\begin{algorithm}
    \DontPrintSemicolon
    
    \SetKwInOut{Input}{input}\SetKwInOut{Output}{output}
    \SetKwComment{Comment}{/* }{ */}
    
    \Input{$\mu \in \mathcal{U}(\Omega)$, $d=dim(\Omega)$, $\Omega = \mathbb{S}^2~\text{or}~\mathbb{R}^3 $ 
    \newline $\phantom{lll\gets}$\includegraphics[height=1em]{fig/comment_emo.png} Discrete measure $\frac{1}{N} \sum \delta_{\mathbf{x}_i}$ 
    \newline $v \in \mathbb{R}^{k \times d}$, $p \in \mathbb{R}^{m \times d}$
    }
    \Output{$c$}
    \BlankLine
    
    $c = \{v, p\} \in \mathbb{R}^{(k + n) \times d}$
    
    \For{$i \gets 1$ \KwTo $N_{iter}$} {
        $\mathcal{V}^c \gets \texttt{Voronoi}(c)$ \;
        $b^c \gets \texttt{computeBarycenter}(\mathcal{V}^c, \mu)$ \;
        $c \gets \{v, b^c_{k+1 \dots m}\}$ \;
    }
    \textbf{return} $c$
    \caption{Lloyd Relaxation}
    \label{alg:lloyd}
\end{algorithm}

\section{Implementation Details}
\label{supp:implementation}

\paragraph{Re-split the test set for TanksAndTemples.}

The test sets in the TanksAndTemples datasets comprise one or two video clips, showcasing parts of the reconstructed scene.
\Cref{fig:tnt_old} visualizes the original test view coverage of \textit{M60} and \textit{Truck}.
Notably, a significant portion of the objects are not covered by the original test cameras.
As motivated in~\Cref{sec:motivation}, we propose a novel split of the test set for all four scenes in the TanksAndTemples dataset which aims at providing a more robust evaluation of different view selection methods.
We first put together all training and test views for a particular scene.
Then, an equal number of test views were selected as in the original test set for each scene using \gls{fvs}.
The distance metric considered during this process encompassed both spatial distance defined in~\Cref{equation:d_euc} and photogrammetric distance defined in~\Cref{equation:d_photo}.
We can observe from~\Cref{fig:tnt_new} that our proposed test set is able to cover the reconstructed scene more uniformly.

\input{fig/supp_tnt_testset}

\paragraph{ActiveNeRF~\cite{Pan2022ActiveNeRF}:}

ActiveNeRF regards information gain as the reduction of uncertainty. Thus, it selects the candidate view that maximizes the information gained at each selection step.
The ActiveNeRF\footnote{\url{https://github.com/LeapLabTHU/ActiveNeRF/tree/main}} is implemented on the backbone of vanilla \gls{nerf}.
We re-implemented it using InstantNGP backbone and coined it Active-InstantNGP.
Direct re-implementation of ActiveNeRF on InstantNGP failed to learn the 3D scene due to the reformulation of the \gls{nerf} framework as well as the training and rendering process. 
We highlighted the learning of the radiance field in our adopted loss function.
Due to disparities in the rendering performance of Active-InstantNGP, we report the rendering performance of InstantNGP trained on the training views selected with Active-InstantNGP.

\paragraph{Density-aware NeRF Ensembles~\cite{Sunderhauf2022DensityAwareEnsembles}:}
We referred to the experiment of the next best view selection in~\cite{Sunderhauf2022DensityAwareEnsembles} and implemented Density-aware NeRF Ensembles on InstantNGP within our \gls{nerf} Director framework.
More specifically for each selection, we trained 5 models on the same training views with different random seed initializations.
The training process of each ensemble model comprises 2000 training steps.
Then, we computed the uncertainty for all remaining training views using these 5 models as described in~\cite{Sunderhauf2022DensityAwareEnsembles}.
We selected the training view with the highest uncertainty each time.

\section{Additional Results}
\label{supp:results}

\paragraph{Runtime cost analysis on the NeRF Synthetic dataset.}

We report the runtime cost result in Table \ref{tab: speed-up} measuring the training time of InstantNGP across 5 scenes of NeRF Synthetic and averaged for 10 runs (with different views). 
For a fixed view budget, our proposed \gls{fvs} reaches the performances of the traditional \gls{rs} significantly faster (up to 4$\times$ Speedup).
\textbf{\begin{table}
\centering
\scriptsize
\begin{tabular}{l|cc|cc|rr}\toprule
\multirow{2}{*}{View \#} &\multicolumn{2}{c|}{FVS} &\multicolumn{2}{c|}{RS} &\multirow{2}{*}{Speedup} \\
\cmidrule{2-5}
&mean &$\sigma$ &mean &$\sigma$ & \\\midrule
50 &0.7 &$\pm$0.26 &2.6 &$\pm$0.26 &4.03 \\ 
100 &1.4 &$\pm$0.43 &2.6 &$\pm$0.30 & 1.96 \\
150 &1.7 &$\pm$0.48 &2.7 &$\pm$0.32 &1.57 \\
\bottomrule
\end{tabular}
\caption{Averaged training time (in minutes) and standard deviation ($\sigma$) comparisons of FVS against RS at the converged quality.}
\label{tab: speed-up}
\end{table}}

\paragraph{Quantitative results of Plenoxels.}

We also provide the quantitative results of Plenoxels'~\cite{Fridovich2022Plenoxels} asymptotic performance of adding more views, in terms of \gls{psnr} and \acrshort{ssim} (\Cref{fig:overview_results_supp}).
We reported the results on 5 scenes of the \gls{nerf} Synthetic dataset and 3 scenes (\textit{M60}, \textit{Playground}, and \textit{Truck}) of the TanksAndTemples dataset for 5 repetitions.
Similar trends in performance, relative to the number of views, can be observed with this alternative backbone.

\input{supp/supp_plenoxels}

\paragraph{Qualitative results of InstantNGP.}

\begin{figure*}[t!]
    \centering
    \begin{tabular}{@{}l@{\,\,}|@{\,\,}c@{\,\,}c@{\,\,}c@{\,\,}c@{\,\,}c@{\,\,}c@{}}
        \rotatebox{90}{\scriptsize \quad\quad\quad\, M60} & \includegraphics[width=0.36\linewidth]{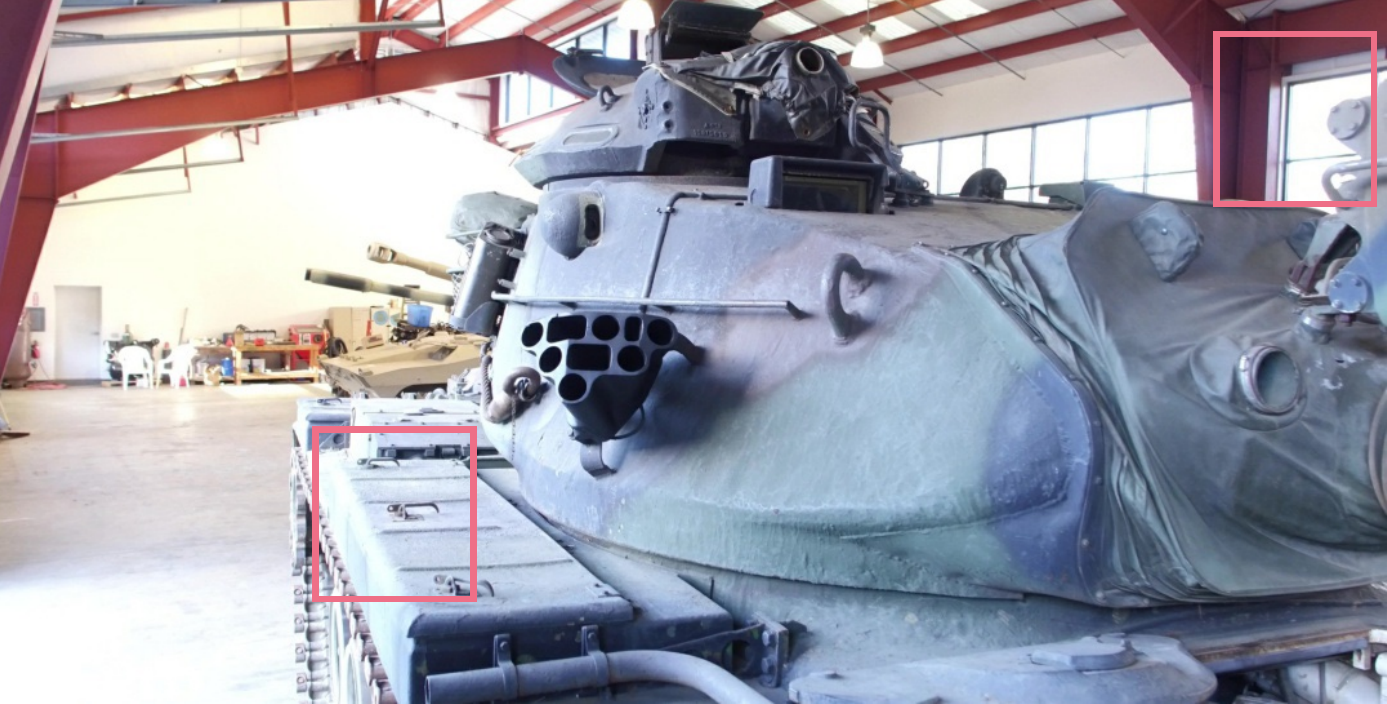} &
        \includegraphics[width=0.1\linewidth]{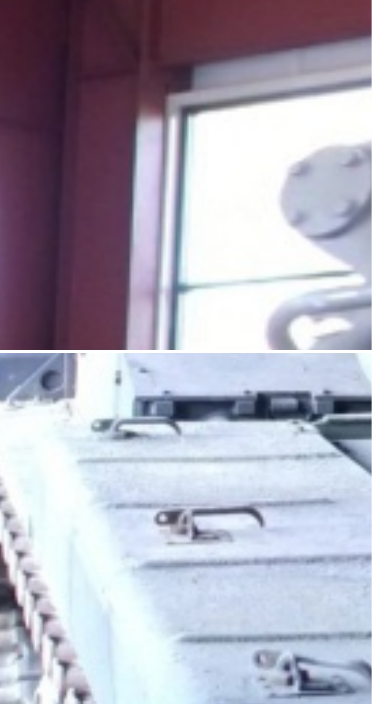} &
        \includegraphics[width=0.1\linewidth]{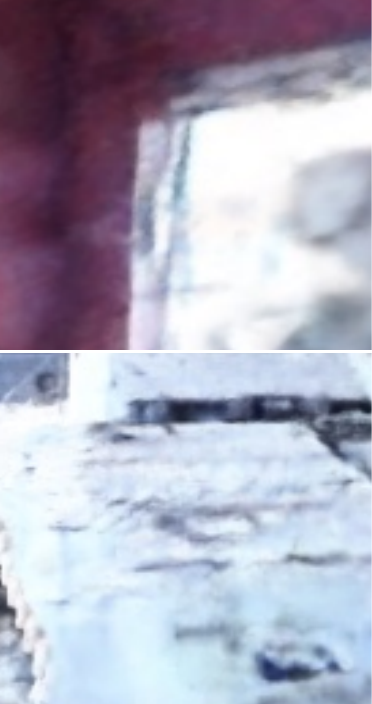} &
        \includegraphics[width=0.1\linewidth]{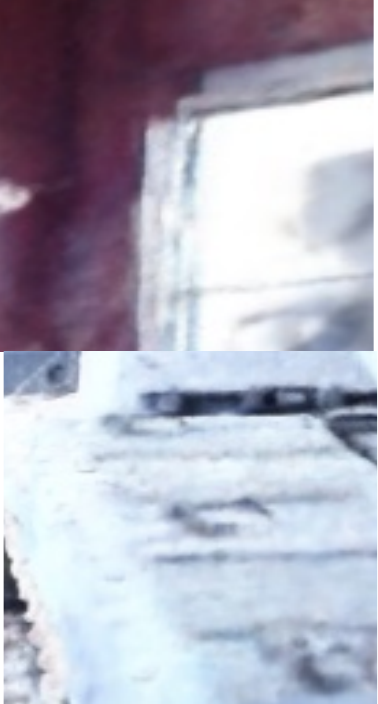} &
        \includegraphics[width=0.1\linewidth]{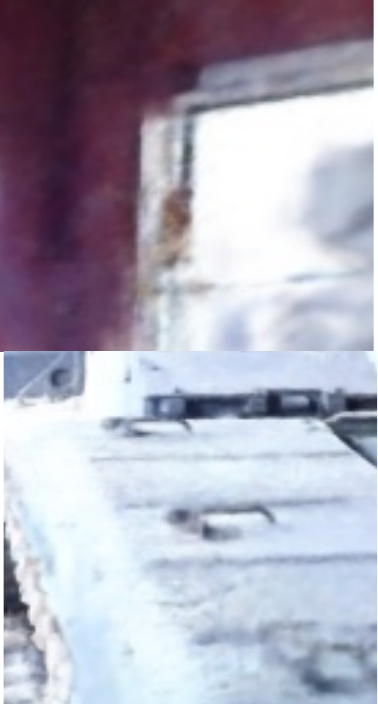} &
        \includegraphics[width=0.1\linewidth]{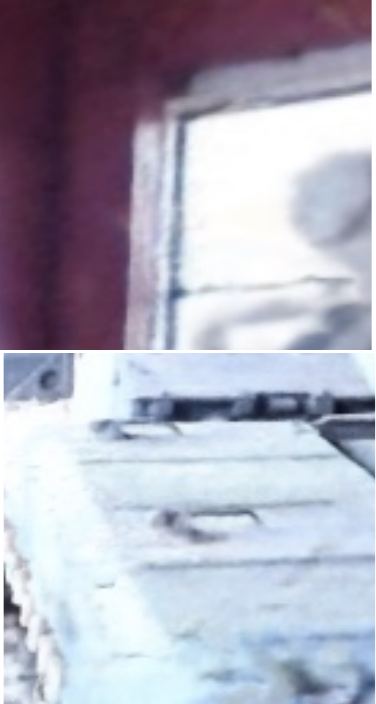} \\
        \rotatebox{90}{\scriptsize \quad\quad\quad\, Playground} & \includegraphics[width=0.36\linewidth]{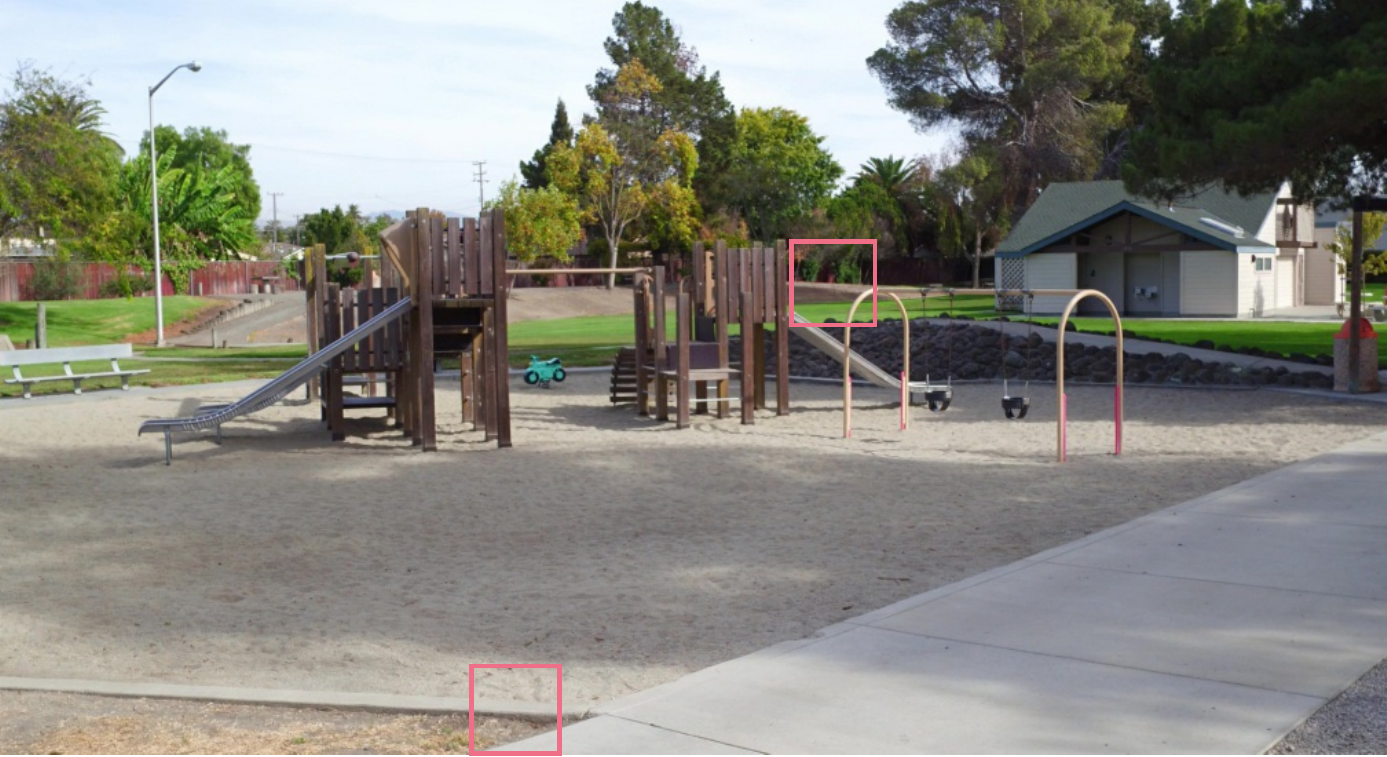} &
        \includegraphics[width=0.1\linewidth]{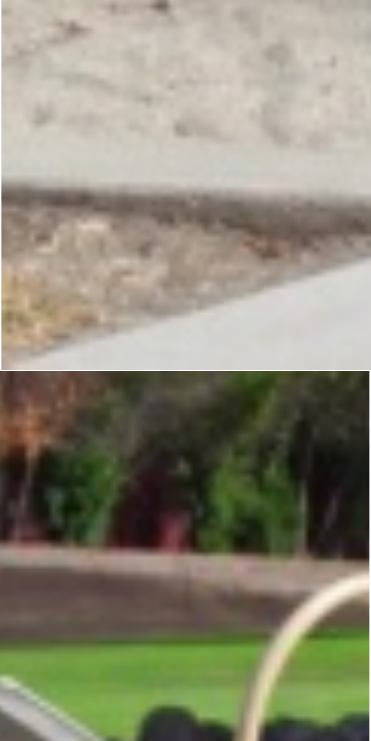} &
        \includegraphics[width=0.1\linewidth]{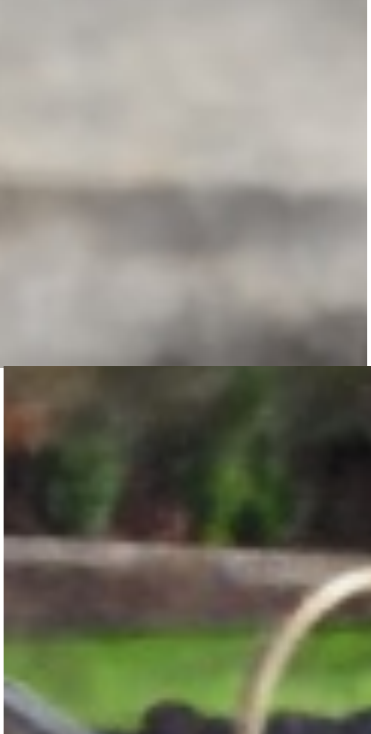} &
        \includegraphics[width=0.1\linewidth]{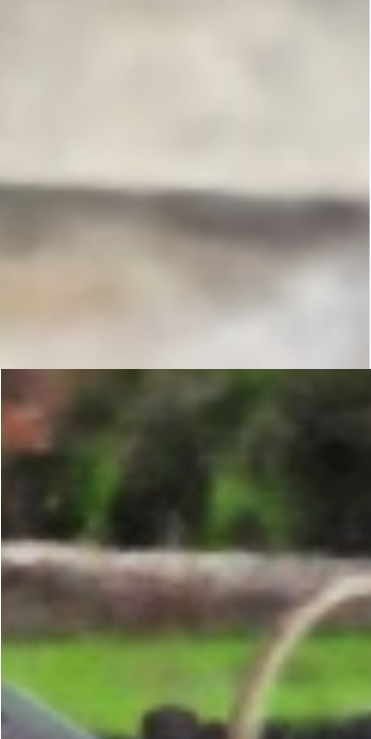} &
        \includegraphics[width=0.1\linewidth]{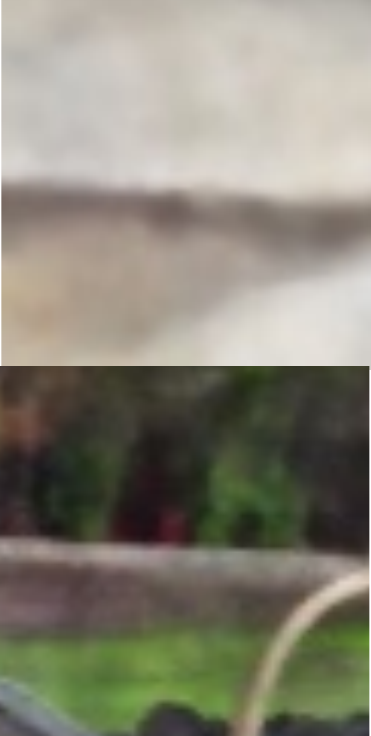} &
        \includegraphics[width=0.1\linewidth]{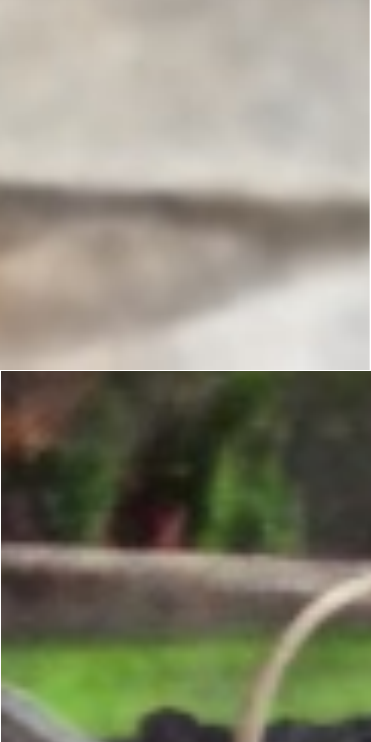} \\
        \rotatebox{90}{\scriptsize \quad\quad\quad\, Train} & \includegraphics[width=0.36\linewidth]{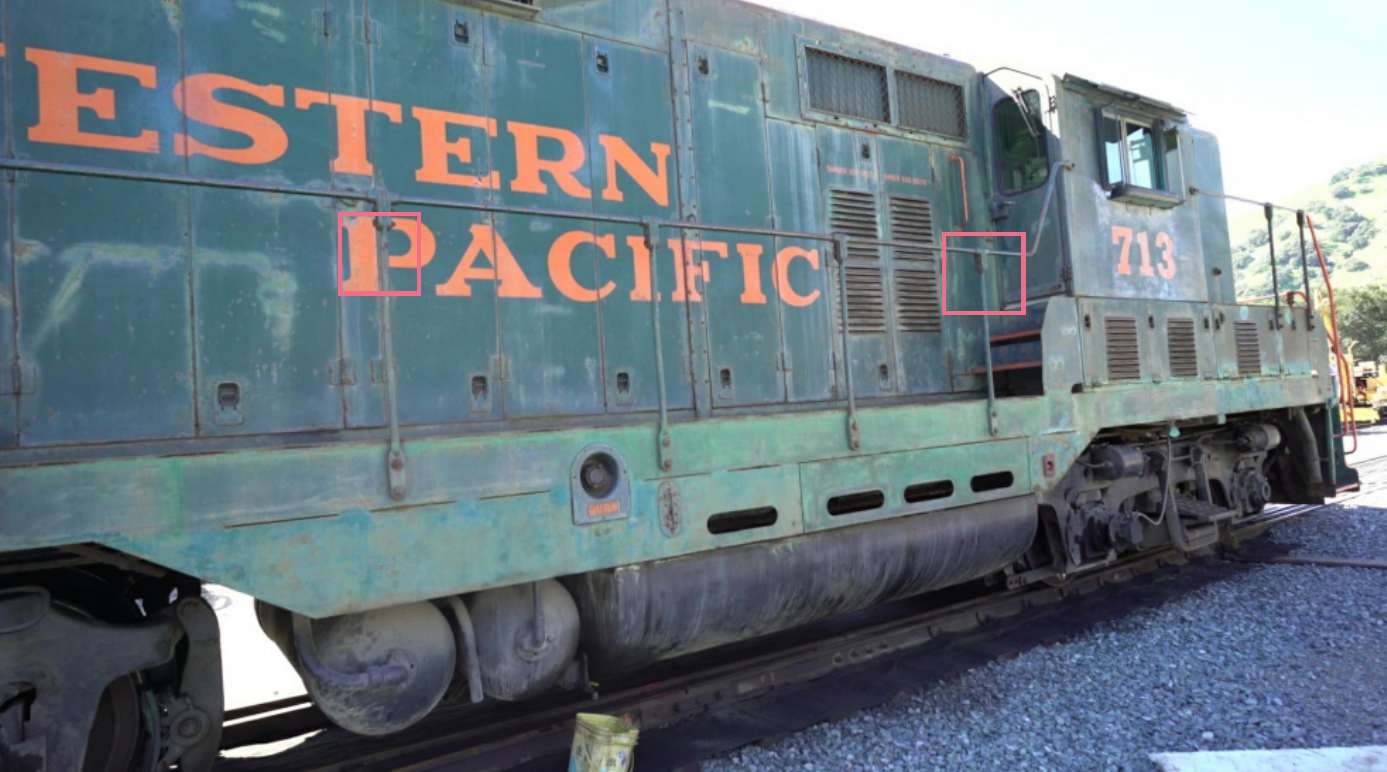} &
        \includegraphics[width=0.1\linewidth]{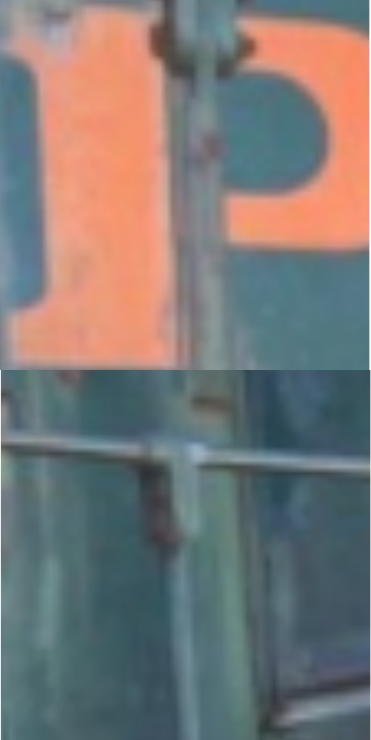} &
        \includegraphics[width=0.1\linewidth]{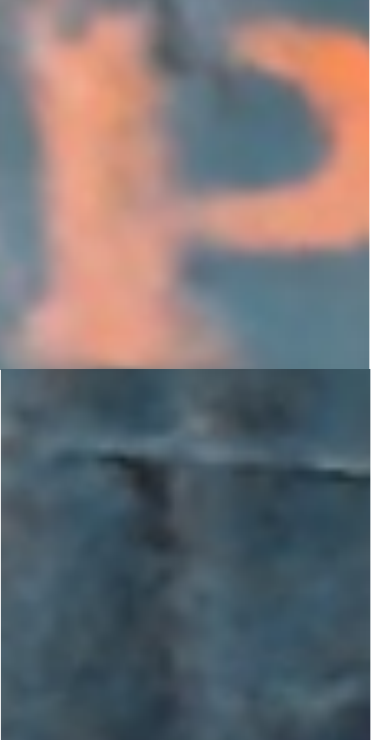} &
        \includegraphics[width=0.1\linewidth]{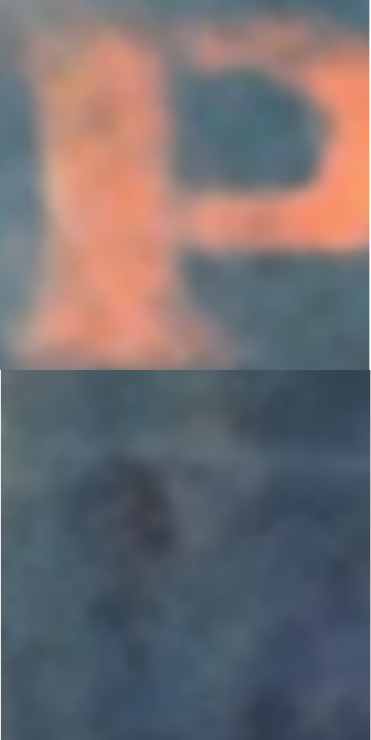} &
        \includegraphics[width=0.1\linewidth]{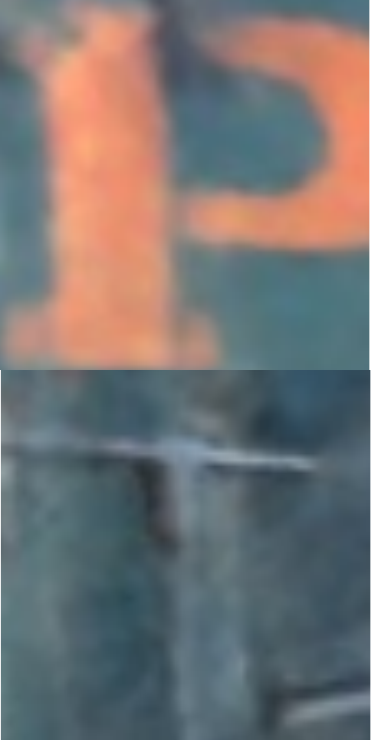} &
        \includegraphics[width=0.1\linewidth]{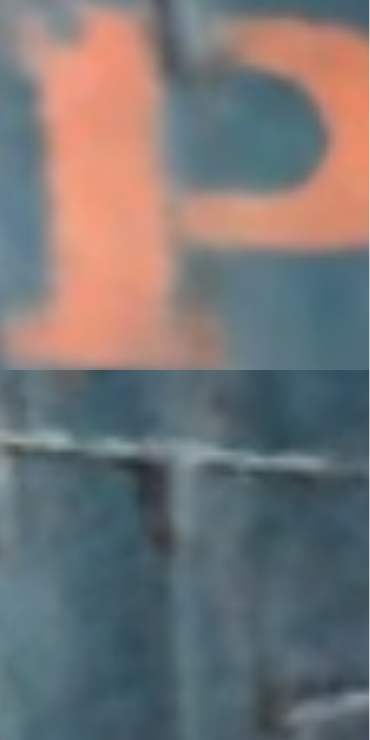} \\
        \rotatebox{90}{\scriptsize \quad\quad\quad\, Truck} & \includegraphics[width=0.36\linewidth]{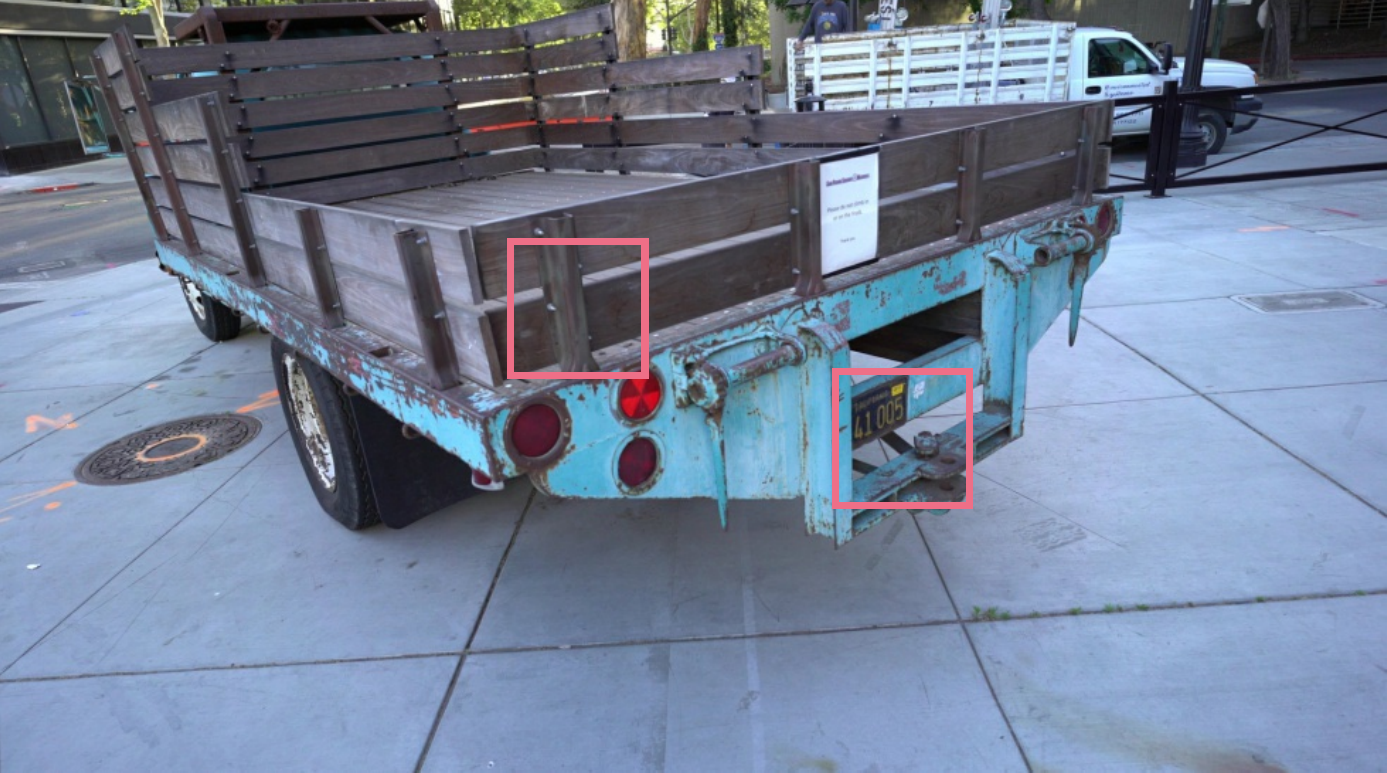} &
        \includegraphics[width=0.1\linewidth]{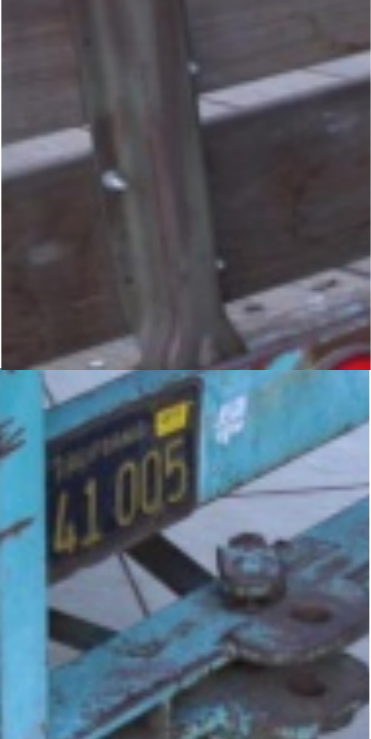} &
        \includegraphics[width=0.1\linewidth]{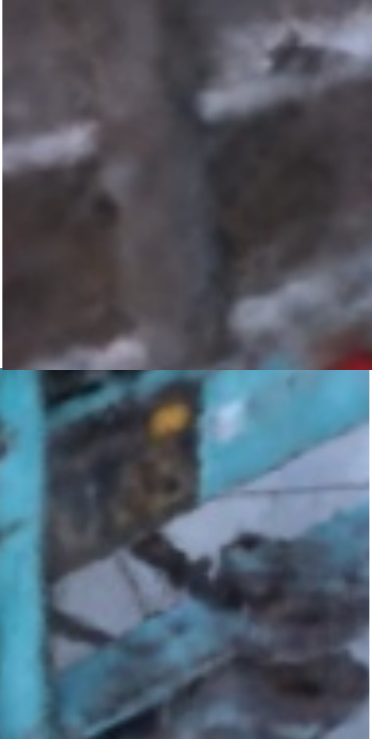} &
        \includegraphics[width=0.1\linewidth]{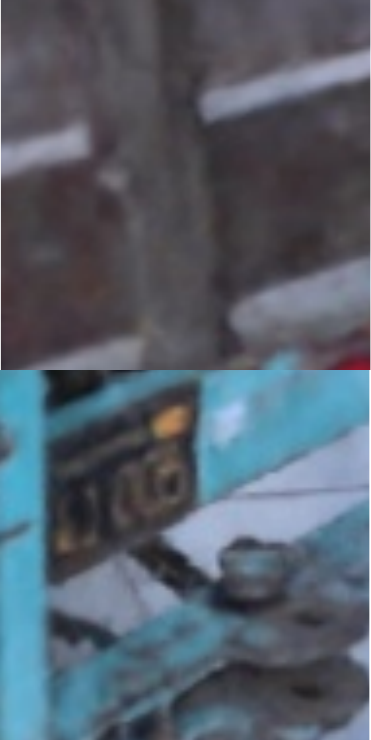} &
        \includegraphics[width=0.1\linewidth]{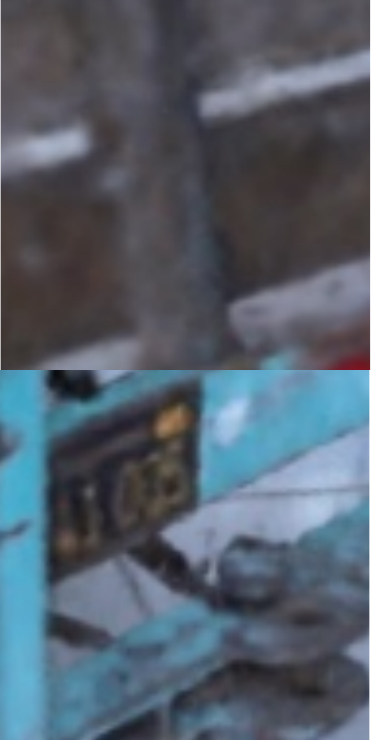} &
        \includegraphics[width=0.1\linewidth]{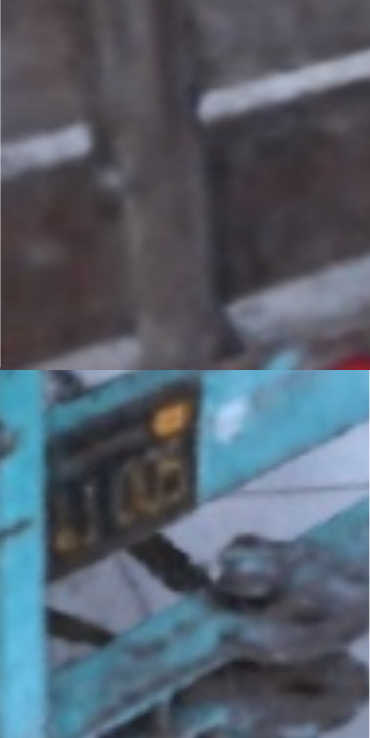} \\
        \hline
        & & \scriptsize \textit{G.T.} & \scriptsize \textit{RS} & \scriptsize \textit{\cite{Sunderhauf2022DensityAwareEnsembles}} & \scriptsize \textit{IGS(vMF)} & \scriptsize \textit{FVS}\\
    \end{tabular}
    \caption{Qualitative comparison results of four view selection methods on the TanksAndTemples dataset with 80 training views.
    }
    \label{fig:tnt_qualitative}
\end{figure*}

\begin{figure*}[t!]
    \centering
    \begin{tabular}{@{}l@{\,\,}|@{\,\,}c@{\,\,}c@{\,\,}c@{\,\,}c@{\,\,}c@{\,\,}c@{}}
        \rotatebox{90}{\scriptsize \quad\quad\quad\, materials} & \includegraphics[width=0.24\linewidth]{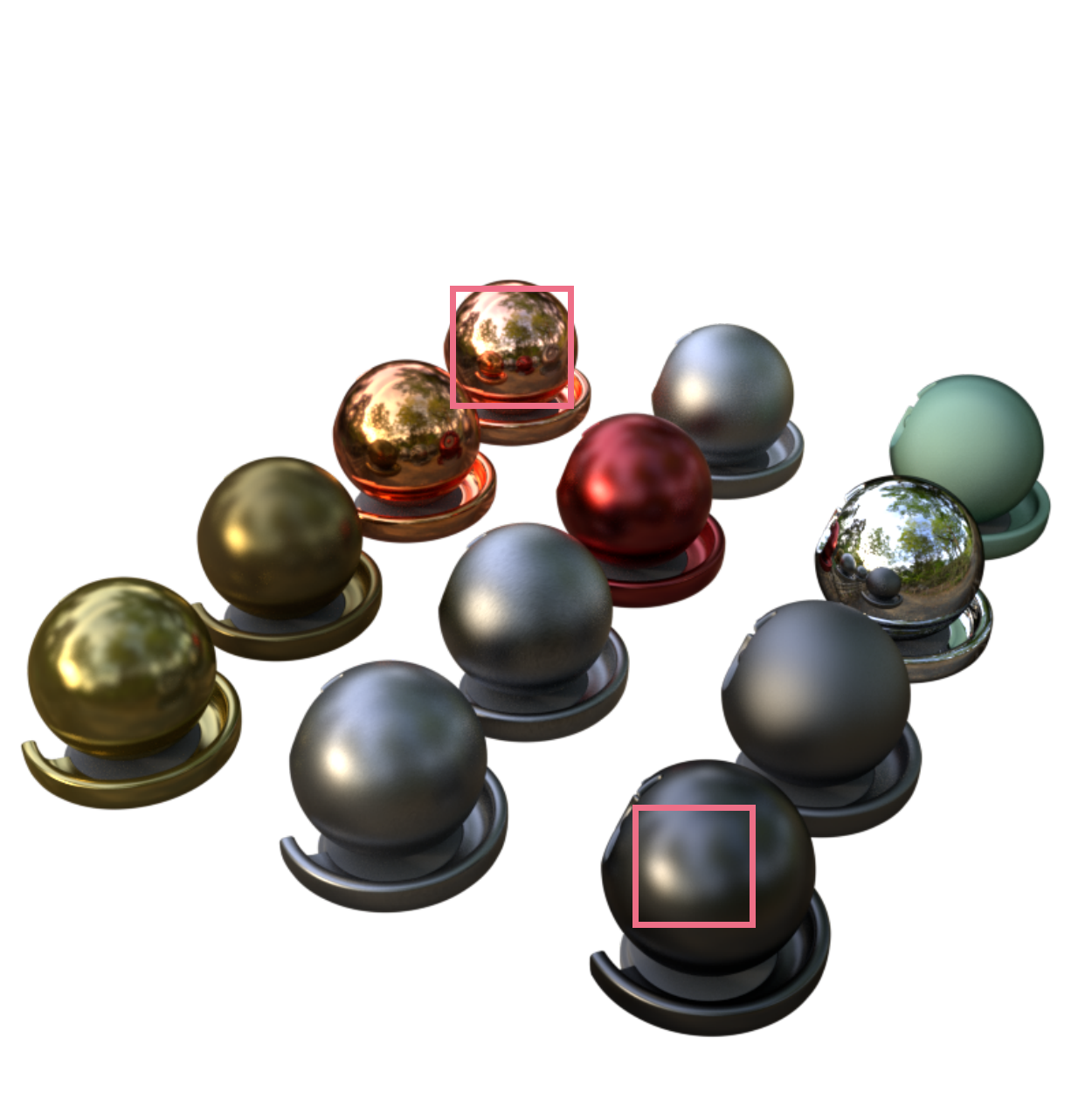} &
        \includegraphics[width=0.09\linewidth]{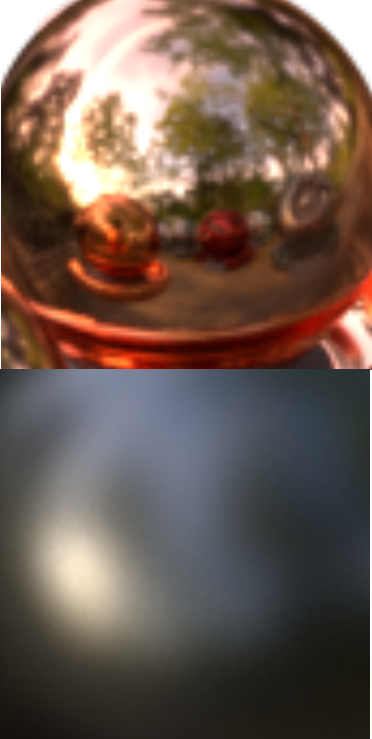} &
        \includegraphics[width=0.09\linewidth]{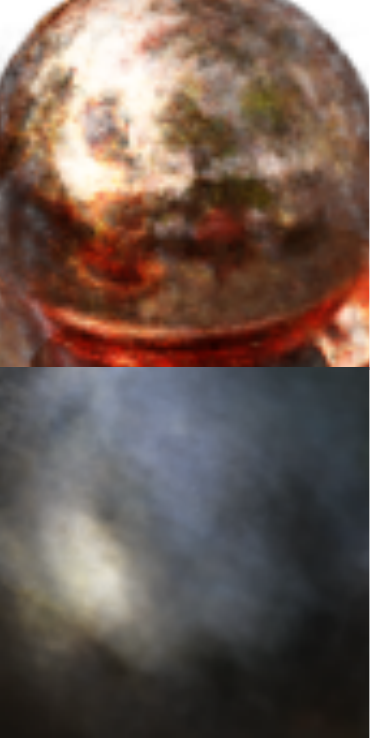} &
        \includegraphics[width=0.09\linewidth]{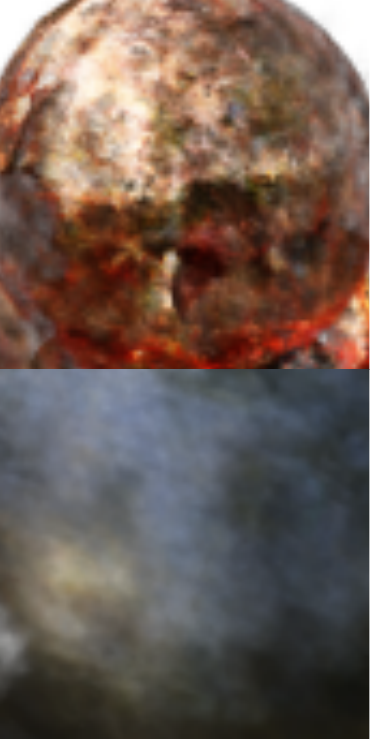} &
        \includegraphics[width=0.09\linewidth]{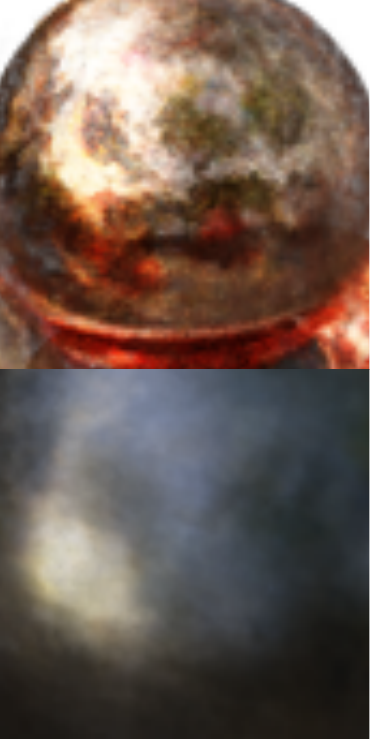} &
        \includegraphics[width=0.09\linewidth]{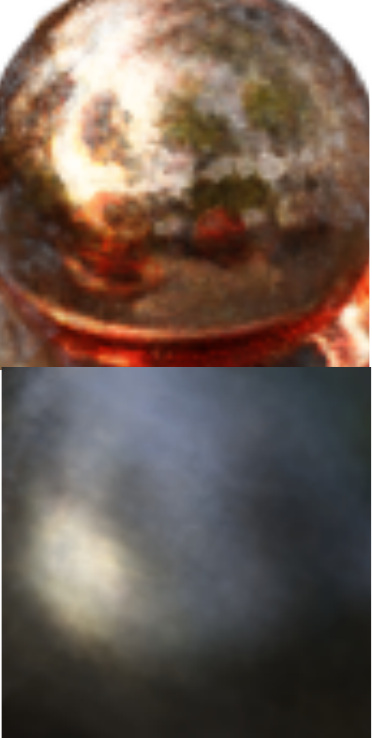} \\
        \rotatebox{90}{\scriptsize \quad\quad\quad\, ficus} & \includegraphics[width=0.24\linewidth]{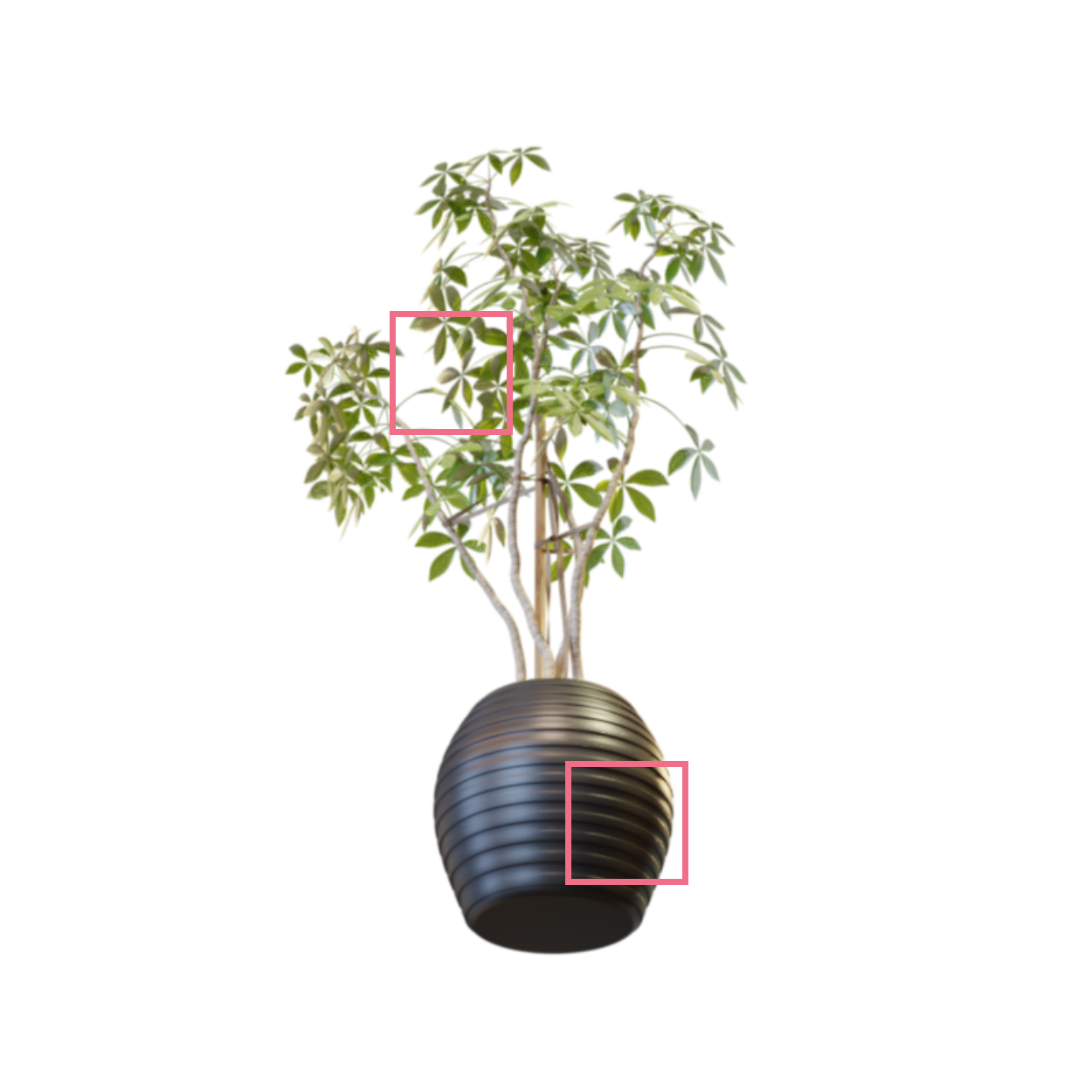} &
        \includegraphics[width=0.09\linewidth]{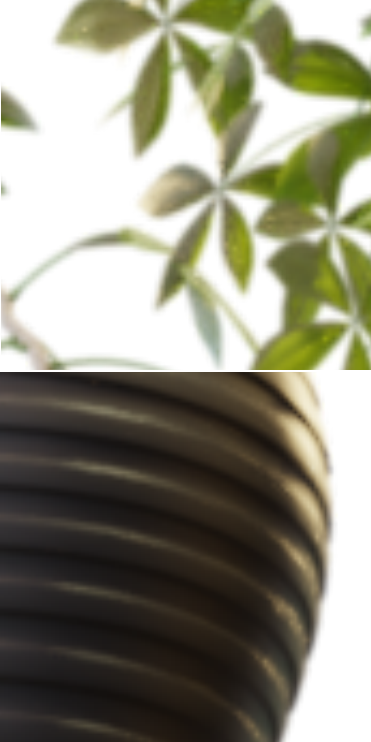} &
        \includegraphics[width=0.09\linewidth]{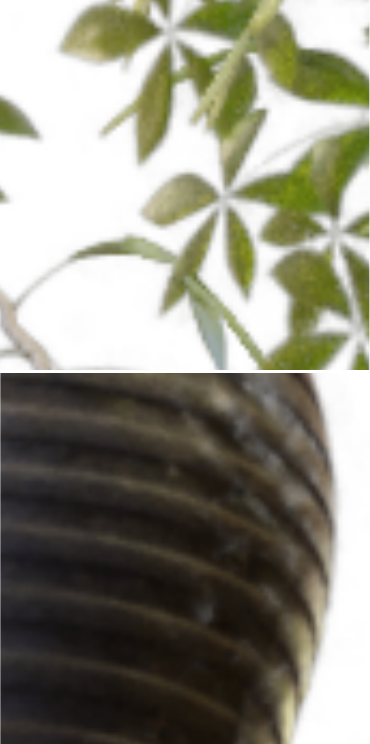} &
        \includegraphics[width=0.09\linewidth]{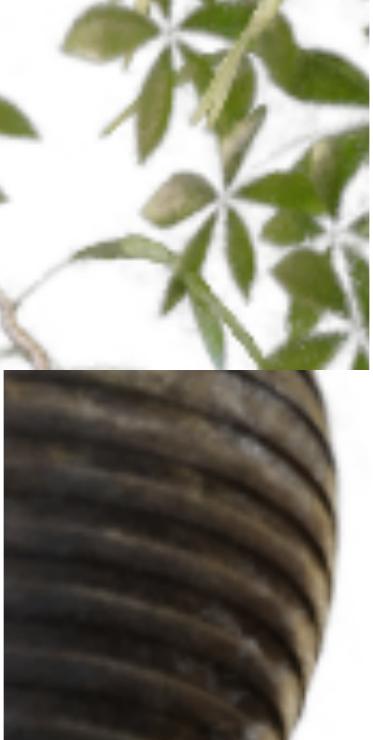} &
        \includegraphics[width=0.09\linewidth]{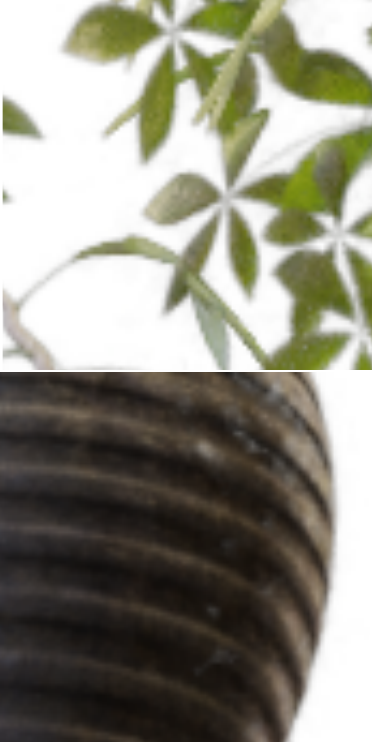} &
        \includegraphics[width=0.09\linewidth]{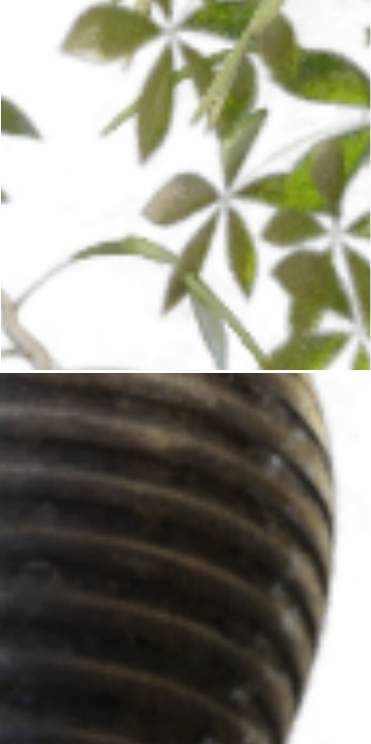} \\
        \rotatebox{90}{\scriptsize \quad\quad\quad\, chair} & \includegraphics[width=0.24\linewidth]{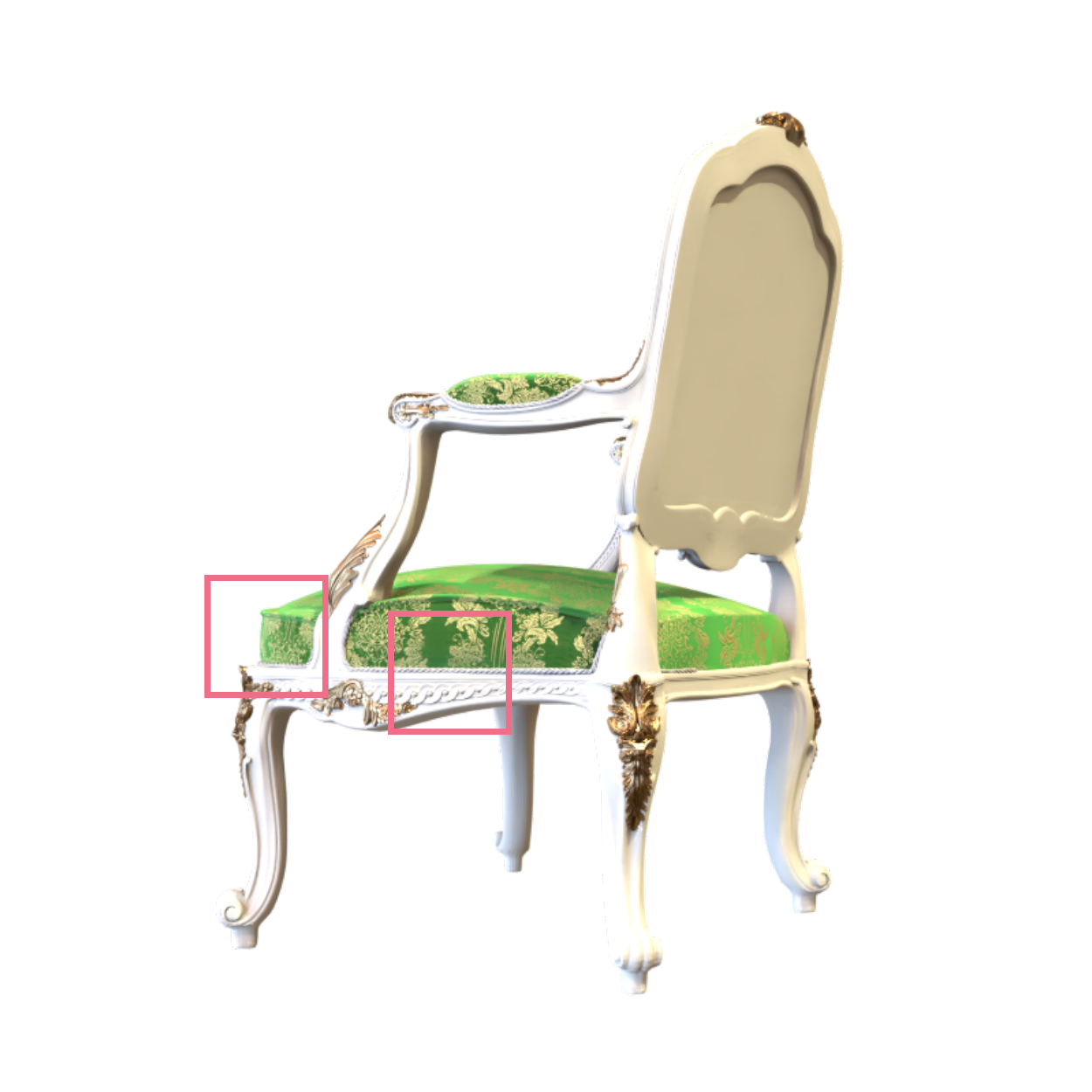} &
        \includegraphics[width=0.09\linewidth]{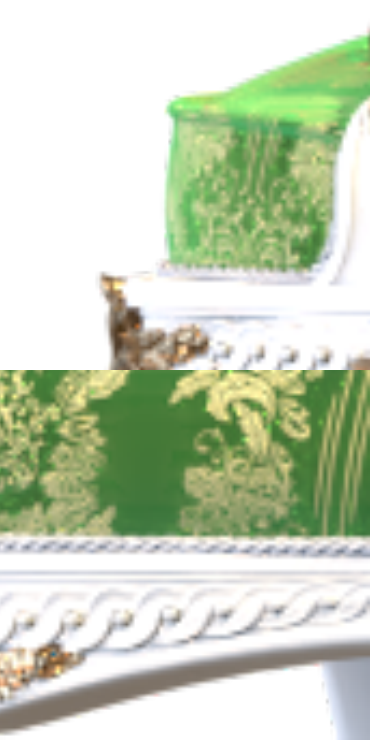} &
        \includegraphics[width=0.09\linewidth]{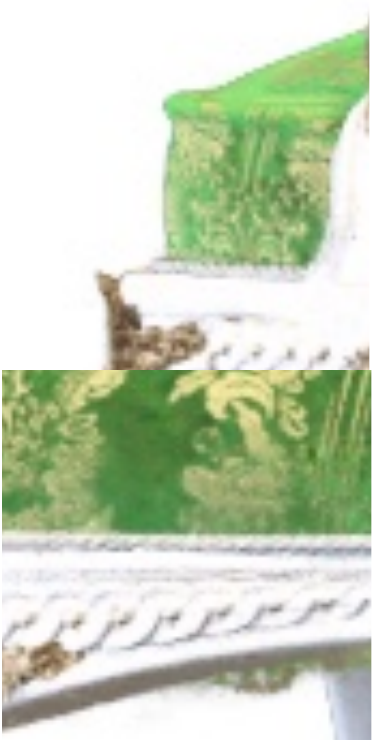} &
        \includegraphics[width=0.09\linewidth]{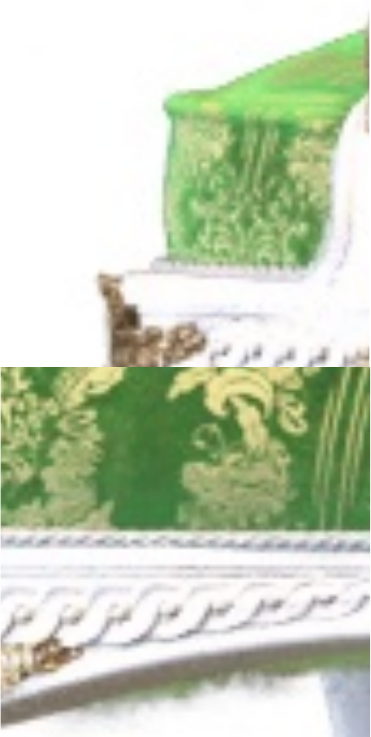} &
        \includegraphics[width=0.09\linewidth]{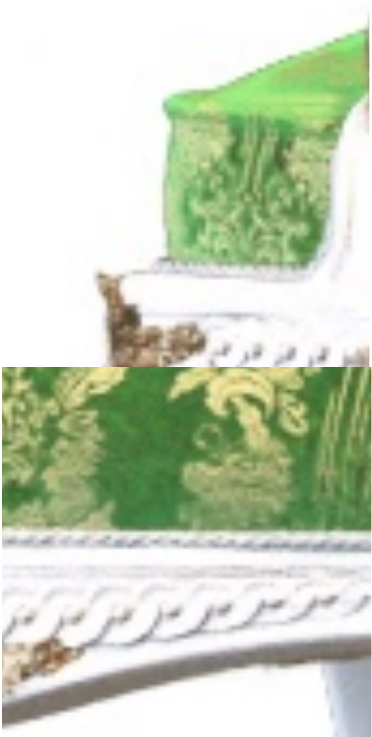} &
        \includegraphics[width=0.09\linewidth]{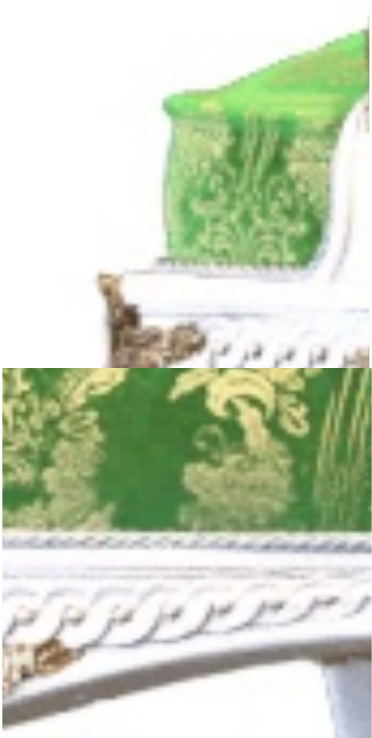} \\
        \hline
        & & \scriptsize \textit{G.T.} & \scriptsize \textit{RS} & \scriptsize \textit{\cite{Sunderhauf2022DensityAwareEnsembles}} & \scriptsize \textit{IGS(vMF)} & \scriptsize \textit{FVS}\\
    \end{tabular}
    \caption{Qualitative comparison results of four view selection methods on the NeRF Synthetic dataset with 80 training views.
    }
    \label{fig:blender_qualitative}
\end{figure*}

We provide the qualitative results of InstantNGP on both the TanksAndTemples and the NeRF Synthetic dataset, as shown in~\Cref{fig:tnt_qualitative} and~\Cref{fig:blender_qualitative} respectively.
We compared our proposed \gls{fvs} and \gls{hs}(vMF) with the baseline \gls{rs} and view selection method in~\cite{Sunderhauf2022DensityAwareEnsembles}.
It can be observed that our proposed methods can generate a clearer and sharper appearance.

{
    \small
    \bibliographystyle{ieeenat_fullname}
    \bibliography{references}
}

%% file: fig/supp_lloyd_motivation.tex
\begin{figure}[h!]
    \centering
    \begin{subfigure}[]{0.488\columnwidth}
        \includegraphics[width=1.\columnwidth]{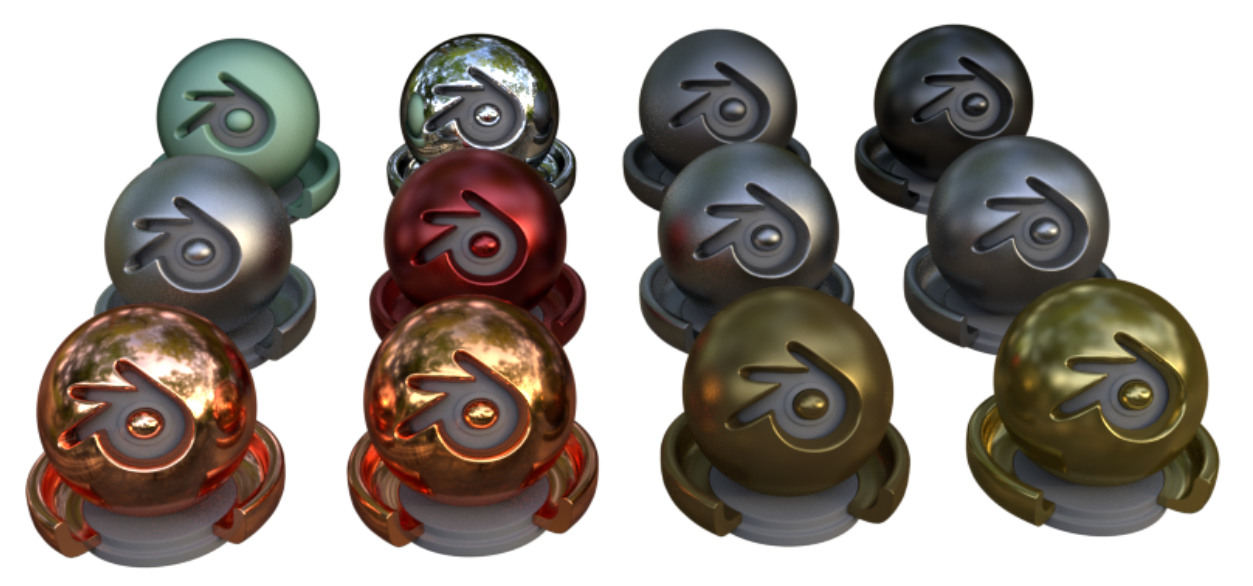}
        \caption{}
        \label{fig:supp_lloyd_GT}
    \end{subfigure}      
    \hfill
    \begin{subfigure}[]{0.488\columnwidth}
        \includegraphics[width=1.\columnwidth]{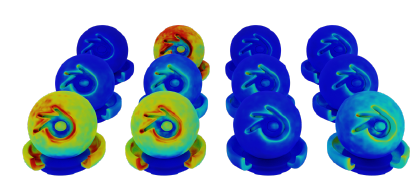}
        \caption{}
        \label{fig:supp_lloyd_error}
    \end{subfigure}
    \caption{Scene \textit{materials} ground truth image (a); Its re-projected PSNR from all rendered images to the mesh (b), where \textit{red} means lower PSNR.}
    \label{fig:supp_lloyd}
\vspace{-8pt}
\end{figure}

%% file: fig/supp_tnt_testset.tex
\begin{figure*}[h!]
    \centering
    \begin{subfigure}[]{1.0\textwidth}
        \includegraphics[width=0.499\textwidth]{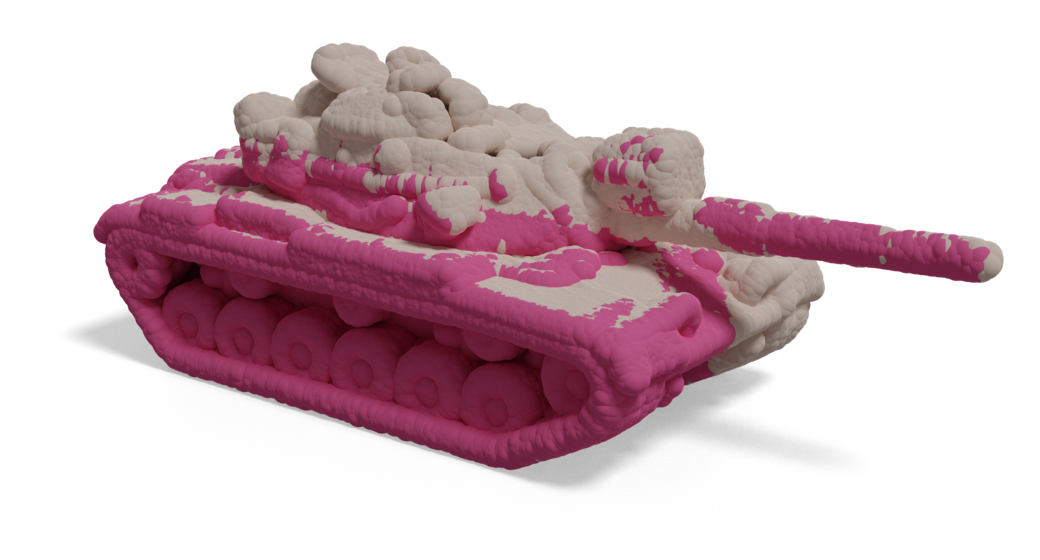}
        \includegraphics[width=0.499\textwidth]{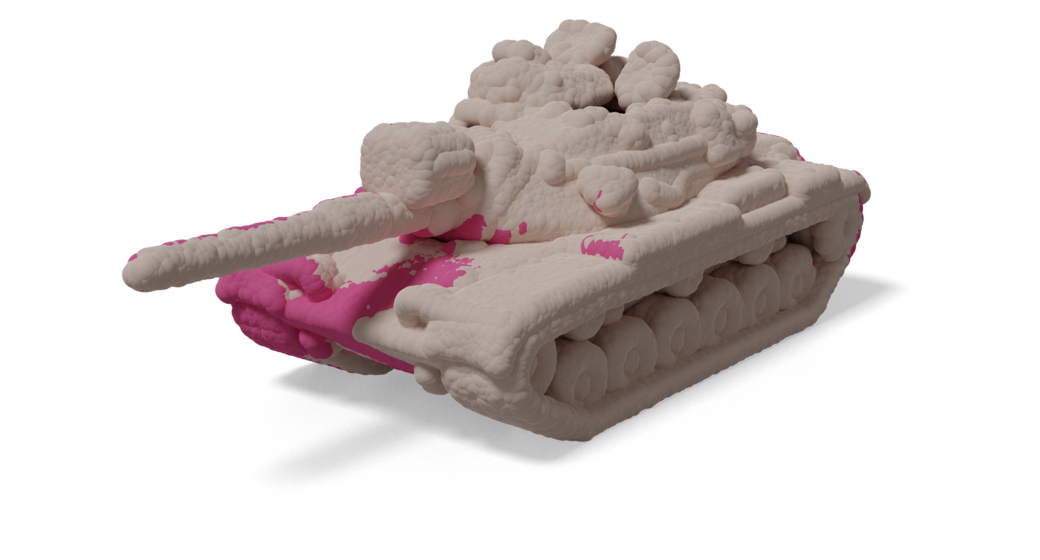}
        \includegraphics[width=0.499\textwidth]{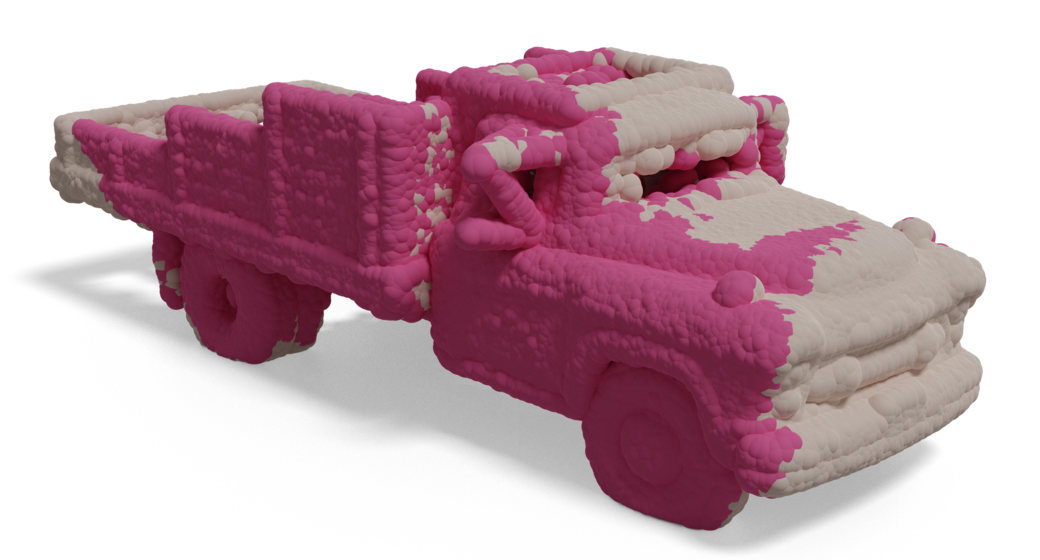}
        \includegraphics[width=0.499\textwidth]{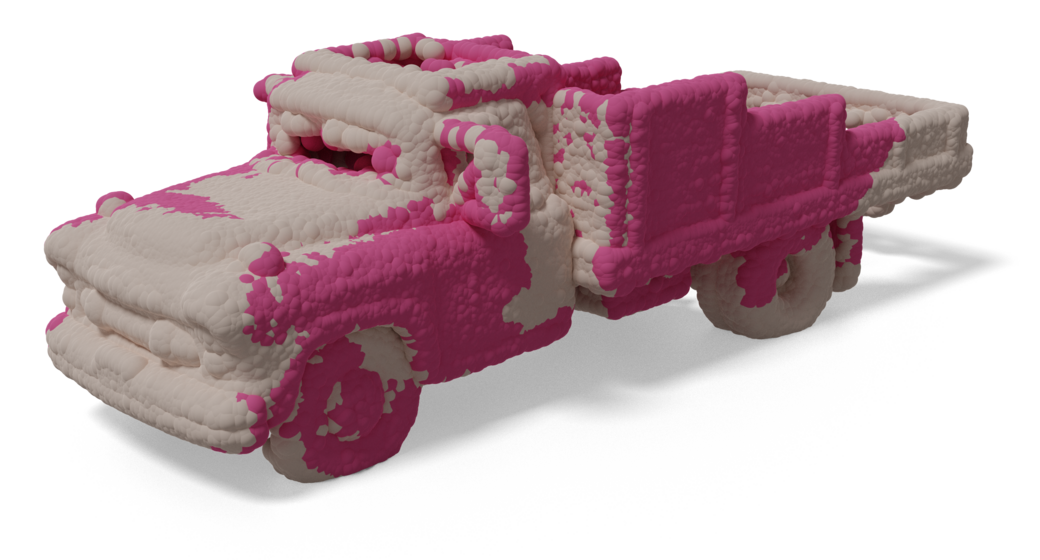}
        \caption{}
        \label{fig:tnt_old}
    \end{subfigure}      
    \hfill
    \begin{subfigure}[]{1.0\textwidth}
        \includegraphics[width=0.499\textwidth]{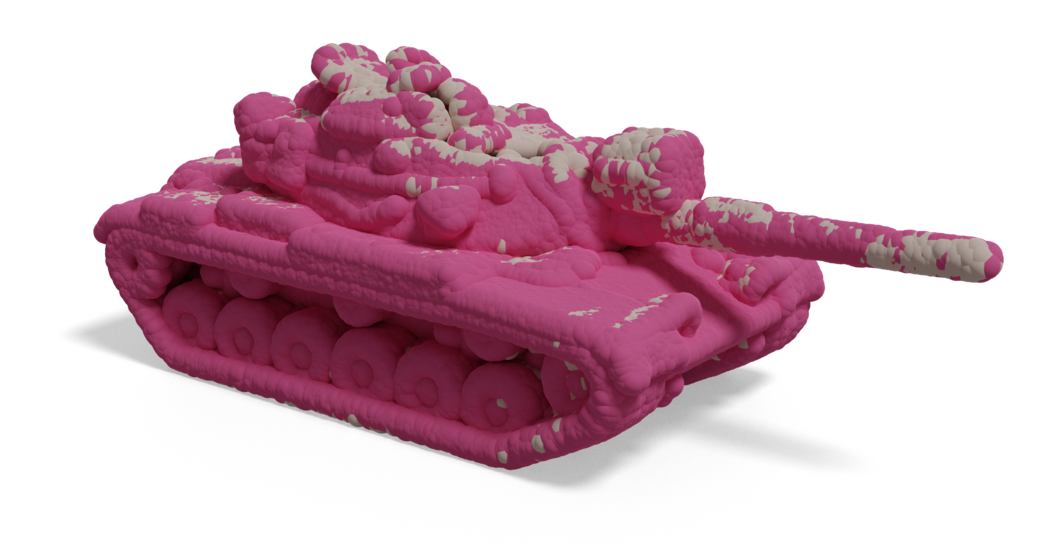}
        \includegraphics[width=0.499\textwidth]{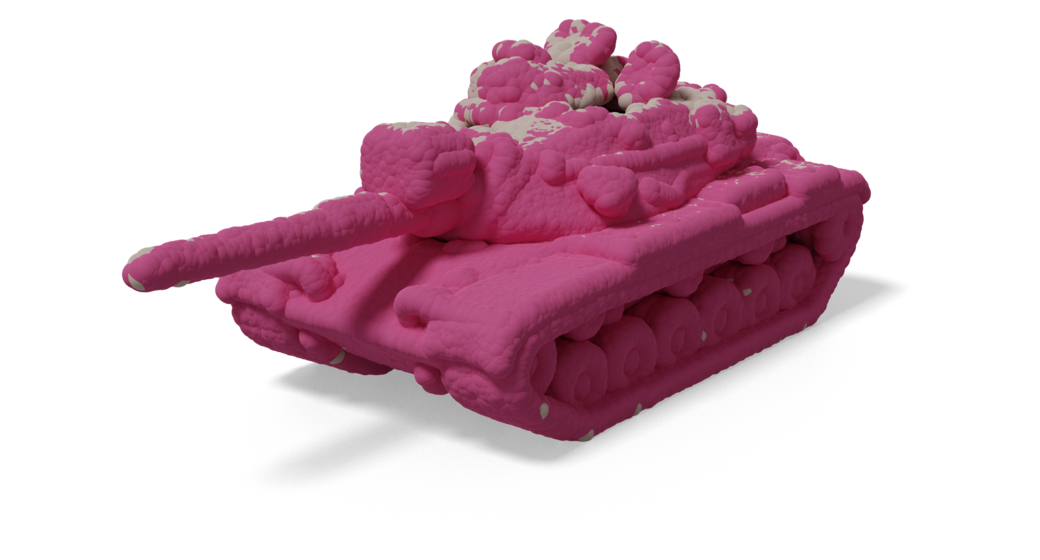}
        \includegraphics[width=0.499\textwidth]{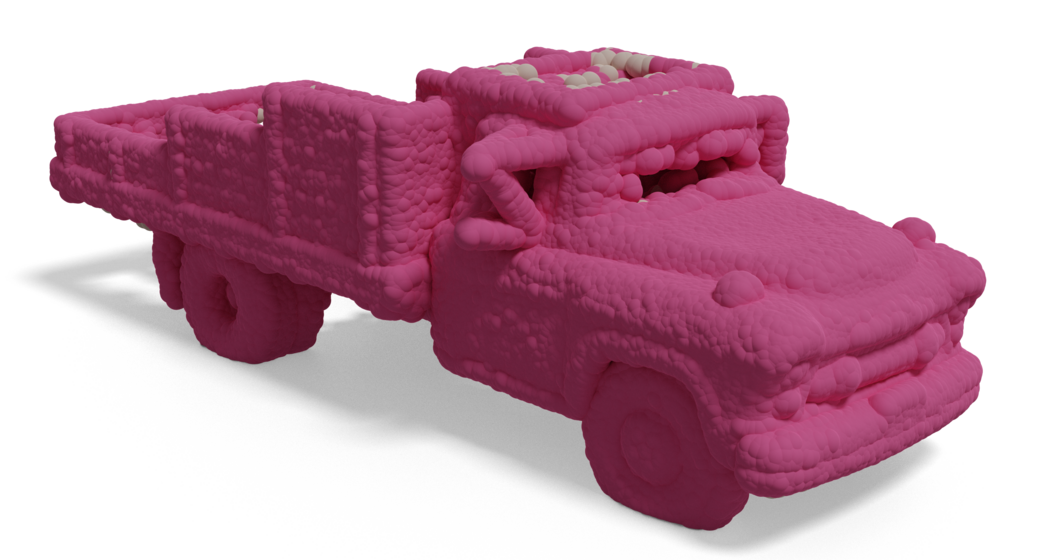}
        \includegraphics[width=0.499\textwidth]{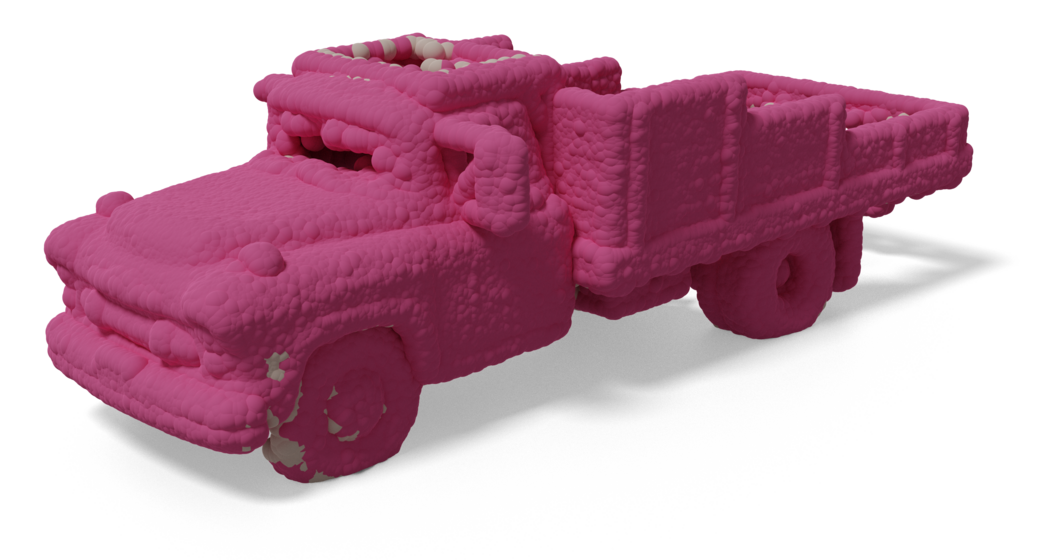}
        \caption{}
        \label{fig:tnt_new}
    \end{subfigure}
    \caption{Visualization of test view coverage on objects \textit{M60} and \textit{Truck}. Pink areas indicate that the ray-mesh intersection is greater than 10 in those regions. (a): Original test view; (b): Proposed test view.}
    \label{fig:supp_tnt}
\vspace{-8pt}
\end{figure*}

%% file: supp/supp_plenoxels.tex
\begin{figure*}[t]
    \centering
    \begin{minipage}{0.9306\textwidth}
        \centering
        \begin{subfigure}[]{0.49\textwidth}
            \centering
            \includegraphics[width=0.99\textwidth]{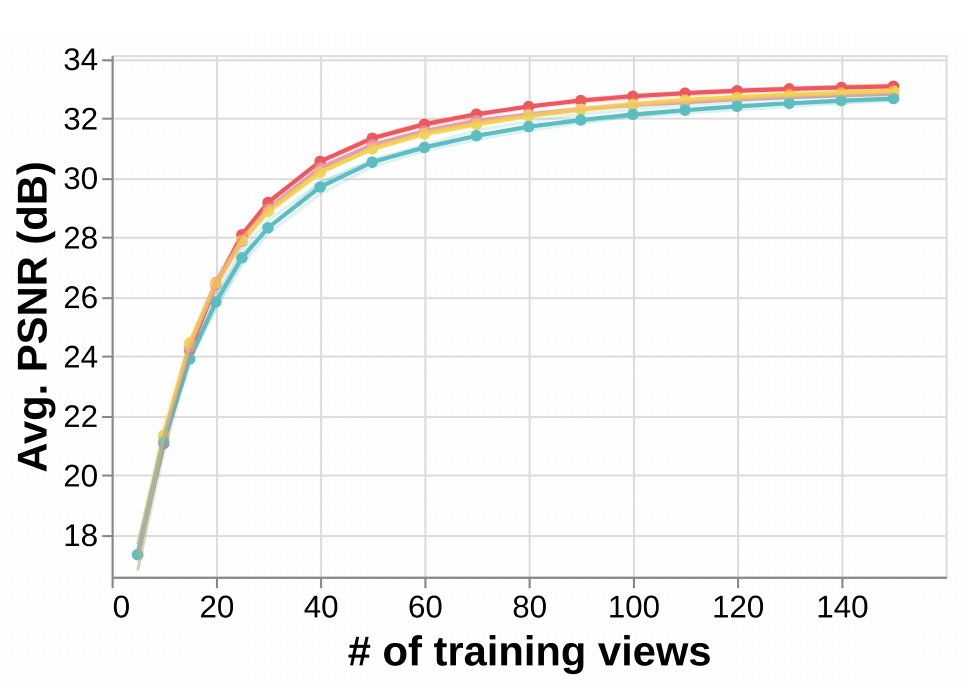}
            \caption{}
            \label{fig:blender_psnr_supp}
        \end{subfigure}
        \hfill
        \begin{subfigure}[]{0.49\textwidth}
            \centering
            \includegraphics[width=0.99\textwidth]{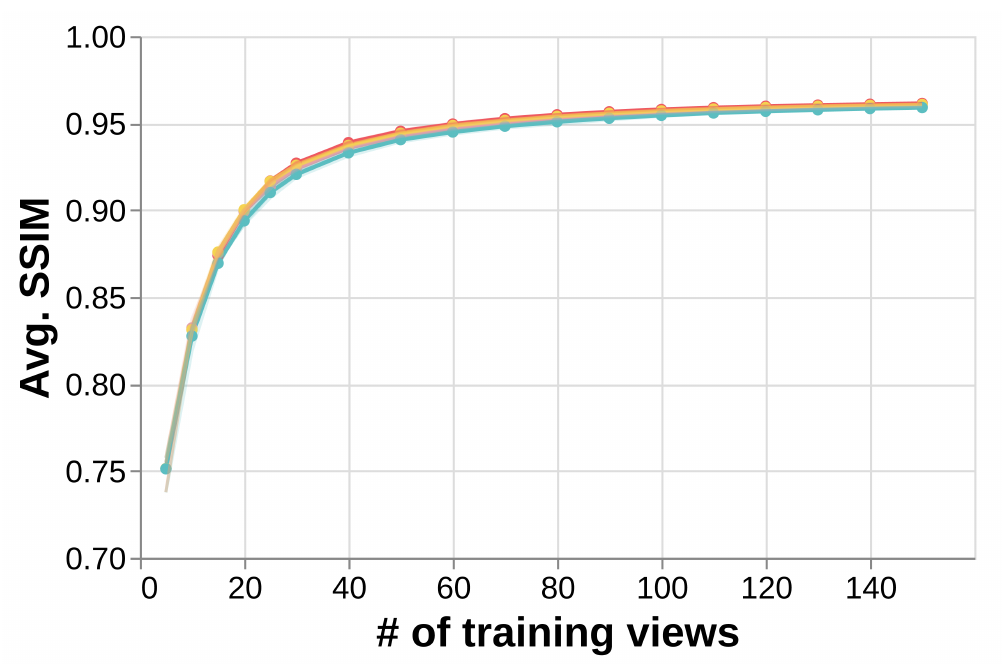}
            \caption{}
            \label{fig:blender_ssim_supp}
        \end{subfigure}
        \hfill
        \begin{subfigure}[]{0.49\textwidth}
            \centering
            \includegraphics[width=0.99\textwidth]{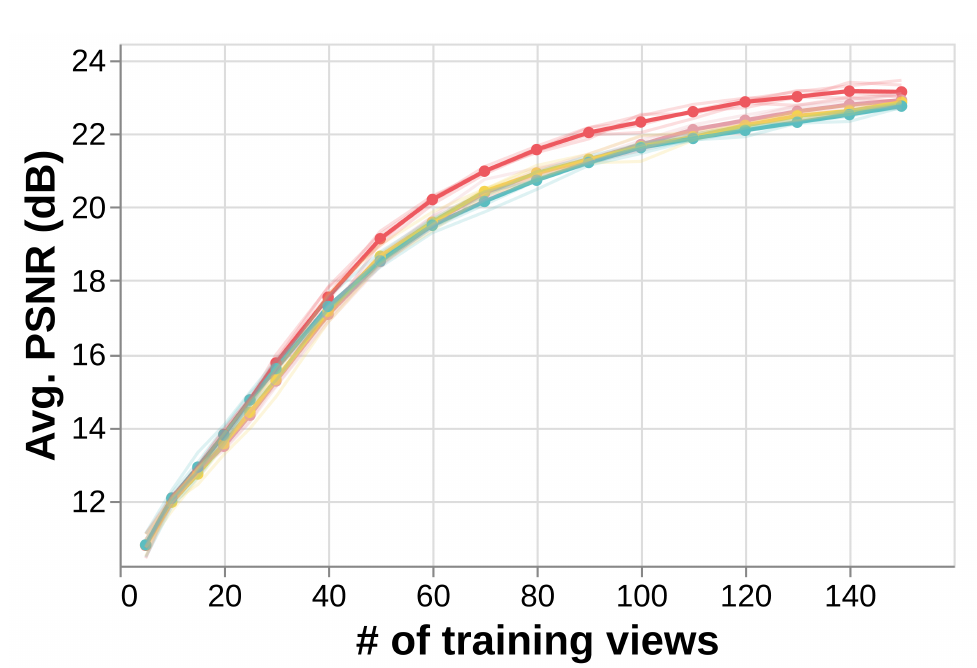}
            \caption{}
            \label{fig:tnt_psnr_supp}
        \end{subfigure}
        \hfill
        \begin{subfigure}[]{0.49\textwidth}
            \centering
            \includegraphics[width=0.99\textwidth]{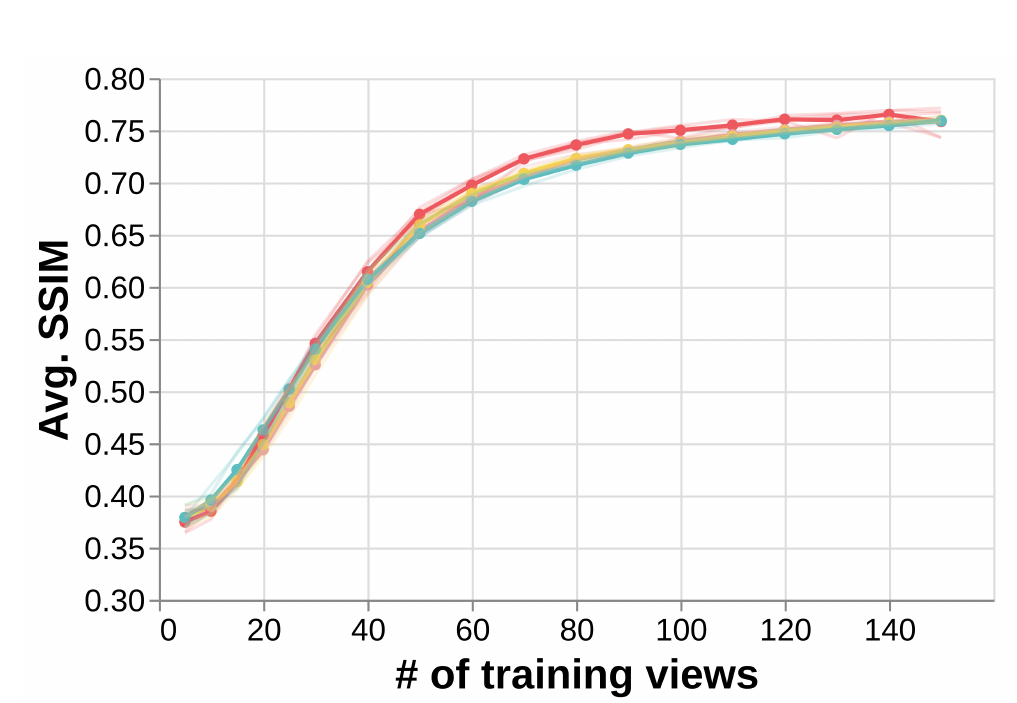}
            \caption{}
            \label{fig:tnt_ssim_supp}
        \end{subfigure}         
    \end{minipage}
    \hfill
    \begin{subfigure}[]{0.056\textwidth}
        \includegraphics[width=1.36\textwidth]{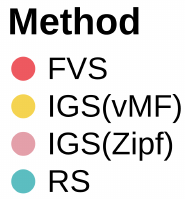}
    \end{subfigure}      
    \vspace{-0pt}
    \caption{Quantitative comparisons of rendering quality on Plenoxels~\cite{Fridovich2022Plenoxels} along with the increase of used training views sampled by different view selection methods. The top row shows the results on the \gls{nerf} Synthetic dataset in terms of PSNR (a) and SSIM (b). The bottom row shows the results on the TanksAndTemples dataset in terms of PSNR (c) and SSIM(d). Low-opacity lines present the results for each repetition, while high-opacity lines present the average result across five repetitions.}
    \label{fig:overview_results_supp}
\vspace{-0pt}
\end{figure*}